\documentclass[final]{cvpr}

\pdfoutput=1

\usepackage{times}
\usepackage{epsfig}
\usepackage{graphicx}
\usepackage{amsmath}
\usepackage{amssymb}
\usepackage{color}
\usepackage{multirow}
\usepackage{tabularx}

\usepackage[pagebackref=true,breaklinks=true,letterpaper=true,colorlinks,bookmarks=false]{hyperref}

\usepackage{cases}
\usepackage{wrapfig}
\usepackage{sidecap}
\usepackage{colortbl}
\usepackage{overpic}
\usepackage{booktabs}
\usepackage{colortbl}
\usepackage{placeins}
\usepackage{mathtools}
\usepackage{algorithmic}
\usepackage{algorithm}
\usepackage{multicol}
\usepackage{float}

\usepackage[resetlabels,labeled]{multibib}
\newcites{A}{Additional References}

\usepackage{rotating}
\usepackage[export]{adjustbox}
\usepackage[table]{xcolor}
\usepackage[inline]{enumitem}

\titleformat{\subsubsection}[runin]{\normalfont\normalsize\bfseries}{\thesubsubsection}{1em}{}[]

\renewcommand{\equref}[1]{Eq.~\eqref{#1}}
\newcommand{\algref}[1]{Alg.~\ref{#1}}

\definecolor{red}{rgb}{0.95,0.4,0.4}
\definecolor{blue}{rgb}{0.4,0.4,0.95}
\definecolor{darkblue}{rgb}{0,0,0.8}
\definecolor{darkred}{rgb}{0.8,0,0}
\definecolor{darkgreen}{rgb}{0,0.5,0}
\definecolor{grey}{rgb}{0.6,0.6,0.6}

\usepackage[normalem]{ulem}

\allowdisplaybreaks

\usepackage{xspace}

\newlength\savewidth

\graphicspath{{figs/matterport_results/}{figs/}{figs/kitti_results/}{figs/supp_figs/matterport_results/}}

\renewcommand*{\ie}{i.e.\@\xspace}
\renewcommand*{\eg}{e.g.\@\xspace}

\usepackage{pifont}

\def\cvprPaperID{4372} 
\def\confYear{CVPR 2021}

\renewcommand{\paragraph}[1]{\noindent\textbf{#1}}

\begin{document}
\title{Panoptic Segmentation Forecasting
\vspace{-0.4cm}}
\author{
Colin Graber$^{1\thanks{\text{Work done during an internship at Niantic.}}}$
~~~
Grace Tsai$^2$
~~~
Michael Firman${}^2$
~~~
Gabriel Brostow${}^{2,3}$
~~~
Alexander Schwing${}^1$\\
${}^1$~University of Illinois at Urbana-Champaign \quad ${}^2$~Niantic \quad ${}^3$~University College London}
\vspace{-0.2cm}
\maketitle
\begin{abstract}
Our goal is to forecast the near future given a set of recent observations. We think this ability to forecast, \ie, to anticipate, is integral for the success of autonomous agents which need not only passively analyze an observation but also must react to it in real-time. Importantly, accurate forecasting hinges upon the chosen scene decomposition. 
We think that superior forecasting can be achieved by decomposing a dynamic scene into individual `things' and background `stuff'. Background `stuff' largely moves because of camera motion, while foreground `things' move because of both camera and individual object motion. 
Following this decomposition, we introduce panoptic segmentation forecasting. 
Panoptic segmentation forecasting opens up a middle-ground between existing extremes, which  either forecast  instance trajectories or  predict the appearance of future image frames. 
To address this task we develop a two-component model: one component learns the dynamics of the background stuff by anticipating odometry, the other one anticipates the dynamics of detected things. 
We establish a leaderboard for this novel task, and validate a state-of-the-art model that outperforms  available  baselines.

\vspace{-0.5cm}
\end{abstract}

\section{Introduction}

An intelligent agent must \emph{anticipate} the outcome of its movement in order to navigate safely~\cite{craik1943,llinas2001}.
Said differently, successful autonomous agents  need to understand the  dynamics of their observations and forecast likely future scenarios in order to successfully operate in an evolving environment. However,  contemporary work in computer vision largely \emph{analyzes} observations, \ie, it studies the apparent. For instance, classical semantic segmentation~\cite{chen2014semantic,long2015fully} aims to delineate the observed outline of objects. While understanding an observation is a first seminal step, it is only part of our job. 
Analyzing the currently observed frame means information is out of date by the time we know the outcome, regardless of the processing time. 
It is even more stale by the time an autonomous agent can perform an action. 
Successful agents therefore need to \emph{anticipate} the future `state' of the observed scene. 
An important question, however, remains open: what is a suitable `state' representation for the future of an observed scene?

\begin{figure}
    \includegraphics[width=\columnwidth]{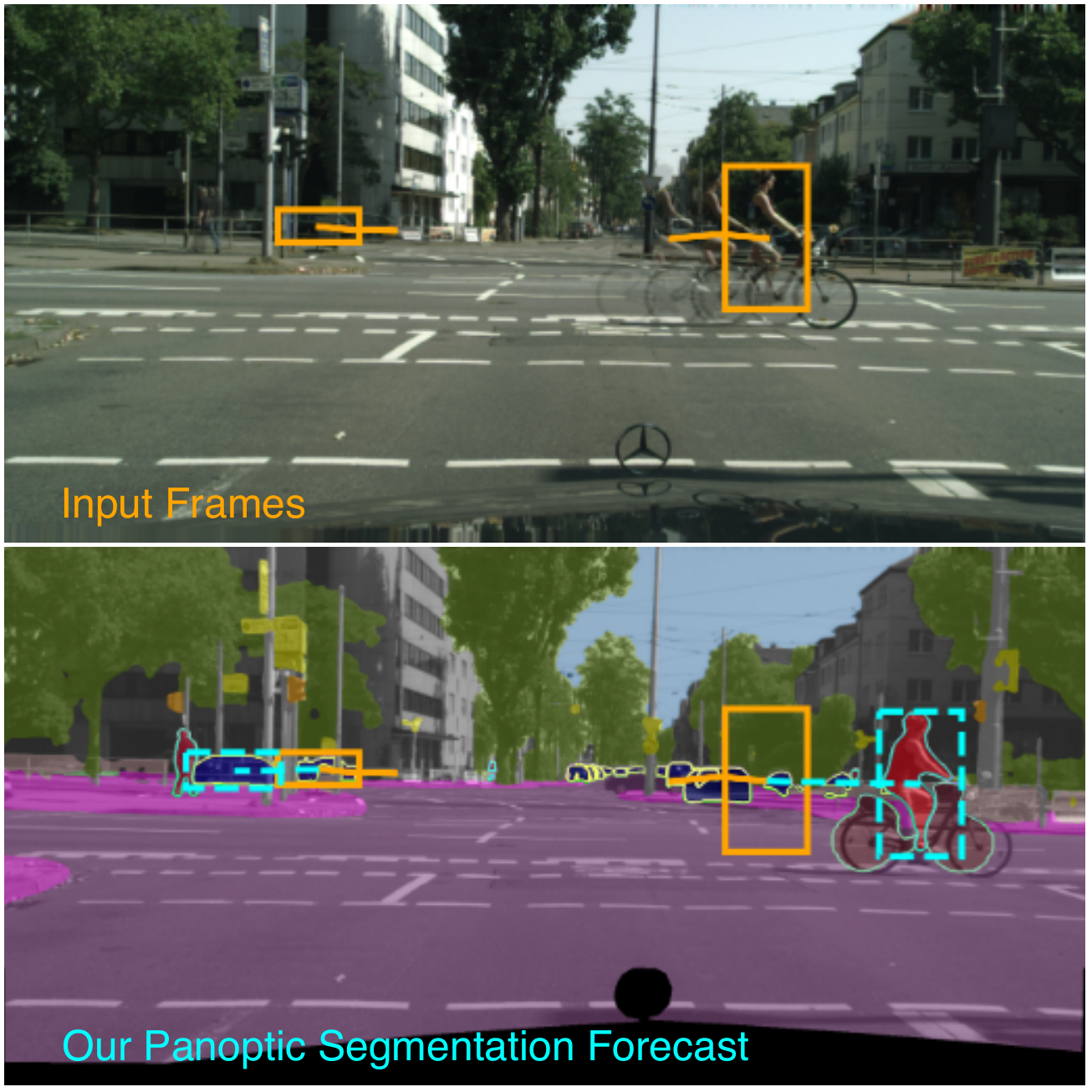}
    \vspace{-0.4cm}
    \caption{We study the novel task of `panoptic segmentation forecasting' and propose a state-of-the-art method that models the motion of individual `thing' instances separately while modeling `stuff' as purely a function of estimated camera motion.}
    \label{fig:teaser}
    \vspace{-0.5cm}
\end{figure}

Panoptic segmentation recently emerged as a rich representation of a scene.
Panoptic segmentation classifies each pixel as either belonging to a foreground instance, the union of which is referred to as `things,' or as a background  class, referred to as `stuff' \cite{heitz2008learning,caesar2018coco}.  
This decomposition is useful for forecasting because we expect different dynamics for each component: `stuff' moves because of the observer's motion, while `things' move because of both observer and object motion. 
Use of panoptic segmentation is further underlined by the fact that it separates different instances of objects, each of which we expect to move individually.

Consequently, we propose to study the novel task of `panoptic segmentation forecasting': 
given a set of observed frames, the goal is to forecast the panoptic segmentation for a set of unobserved frames (Fig.~\ref{fig:teaser}). 
We also propose a first approach to forecasting future panoptic segmentations. 
In contrast to typical semantic forecasting \cite{luc2017predicting, vsaric2019single}, we model the motion of individual object instances and the background separately. 
This makes instance information persistent during forecasting, and allows us to  understand the motion of each moving object.

To the best of our knowledge, we are the first to forecast panoptic segmentations for future, \emph{unseen} frames in an image sequence.
We establish a leaderboard for this task on the challenging Cityscapes dataset~\cite{cordts2016cityscapes} and include a set of baseline algorithms.
Our method for future panoptic segmentation relies on a number of innovations (Sec.~\ref{sec:Method}), that we ablate to prove their value. 
Our method also results in state-of-the-art on previously established tasks of future semantic and instance segmentation. Code implementing models and experiments can be found at \url{https://github.com/nianticlabs/panoptic-forecasting}.
\begin{figure*}[t]
\centering
    \includegraphics[width=\textwidth]{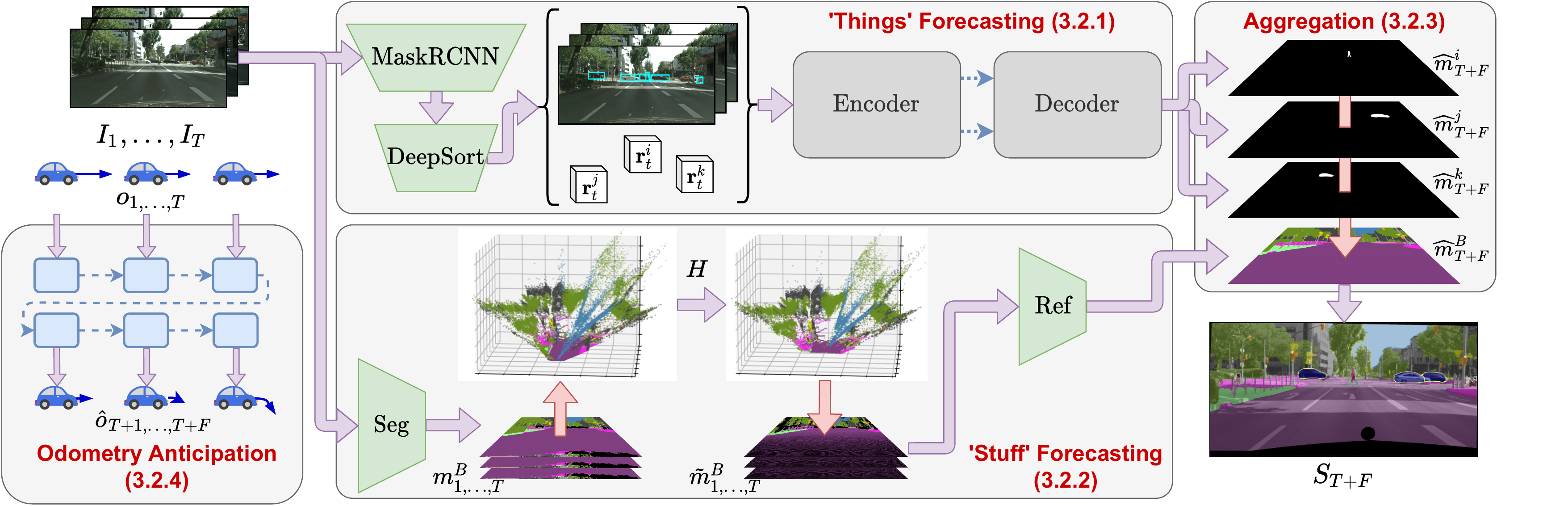}
    \vspace{-0.5cm}
    \caption{\textbf{Method overview.} Given input frames $I_{1, \dots, T}$, our method anticipates the panoptic segmentation $S_{T+F}$ of unseen frame $I_{T+F}$. Our method decomposes the scene into `things' and `stuff' forecasting. `Things' are found via instance segmentation/tracking on the input frames, after which we forecast the segmentation mask and depth of each individual instance (\secref{sec:fg}). Next, `Stuff' is modeled by warping input frame semantics to frame ${T+F}$ using a 3d rigid-body transformation and then passing the result through a refinement model (\secref{sec:bg}). Finally, we aggregate the forecasts from `things' and `stuff' into the final panoptic segmentation $S_{T+F}$ (\secref{sec:agg}). Various components require future odometry $\widehat{o}_{T+1, \dots,T+F}$, which we anticipate using input odometry $o_{1,\dots,T}$ (\secref{sec:odom}). }
    \label{fig:overview}
    \vspace{-0.4cm}
\end{figure*}

\section{Related Work}

We  briefly review work which \emph{analyzes} a single, given frame. 
We then discuss work which \emph{anticipates} info about future, unseen frames. 
To reduce ambiguity, we avoid use of the word 
`predict,' instead using analyze (looking at a current image) or anticipate (hypothesizing for a future frame).

\subsection{Methods That Analyze}

\noindent\textbf{Semantic segmentation:} 
Semantic segmentation has received a considerable amount of attention over decades. The task requires methods to delineate the outline of objects in a given image, either per instance or per object class \cite{sharon2001segmentation, schroff2008object, shotton2008semantic}.
Recently, deep-net-based methods report state-of-the-art results~\cite{long2015fully,badrinarayanan2017segnet,liu2019auto}. 
Many architecture improvements like dilated convolutions \cite{yu2015multi}, skip-connections \cite{ronneberger2015u}, \etc, have  been developed for semantic segmentation before they found use in other tasks. 
Our work differs as we care about \emph{panoptic} segmentation, and we aim to anticipate the segmentation of future, unseen frames.

\noindent\textbf{Panoptic segmentation:} 
Recently, panoptic segmentation~\cite{kirillov2019panoptic, kendall2018multi} has emerged as a generalization of both semantic and instance segmentation. 
It requires methods to give both a per-pixel semantic segmentation of an input image while also grouping pixels corresponding to each object instance. 
This `things' \vs `stuff' view of the world~\cite{heitz2008learning}  comes with its own set of metrics. 
Performing both tasks jointly has the benefit of reducing computation \cite{kirillov2019panoptic,xiong2019upsnet} and enables both tasks to help each other \cite{li2018learning, li2019attention}.
This is similar in spirit to 
multi-task learning \cite{kokkinos2017ubernet,sener2018multi}.
Other works have relaxed the high labeling demands of panoptic segmentation \cite{li2018weakly} or improved  architectures  \cite{Porzi_2019_CVPR, Cheng2020panoptic-deeplab}.
Panoptic segmentation has been extended to videos \cite{kim2020video}, but, again in contrast to our work only analyzing frames available at test time without anticipating future results.

\subsection{Methods That Anticipate}
Anticipating, or synonymously `forecasting,' has received a considerable amount of attention in different communities~\cite{Valassakis2018}. Below, we briefly discuss work on forecasting non-semantic information such as object location before discussing forecasting of semantics and instances.

\noindent\textbf{Forecasting of non-semantic targets:} 
The most common forecasting techniques operate on trajectories. They track and anticipate the future position of individual objects, either in 2D or 3D \cite{dai2020self, martinez2017human, ehrhardt2020relate,YehCVPR2019}. 
For instance, Hsieh \etal~\cite{hsieh2018learning} disentangle position and pose of multiple moving objects -- but only on synthetic data.
Like ours, Kosiorek~\etal~\cite{kosiorek2018sequential} track instances to forecast their future, but only in limited experimental scenarios.
 
Several methods forecast future RGB frames~\cite{liang2017dual,gao2019disentangling, ye2019compositional}. 
Due to the high-dimensional space of the forecasts and due to ambiguity, results can be blurry, despite significant recent advances.
Uncertainty over future frames can be modelled, \eg, using latent variables~\cite{walker2016uncertain, ye2019compositional}.
Related to our approach, Wu \etal~\cite{wu2020future} treat foreground and background separately for RGB forecasting, but they do not model egomotion.
All these methods differ from ours in output and architecture.

\noindent\textbf{Forecasting semantics:} 
Recently, various methods have been proposed to estimate semantics for future, unobserved frames.  
Luc \etal~\cite{luc2017predicting} use a conv net to estimate the future semantics given as input  the current RGB and semantics, while Nabavi \etal~\cite{rochan2018future} use recurrent models with semantic maps as input. 
Chiu \etal~\cite{chiu2020segmenting} further use a teacher net to provide the supervision signal during training, while \v{S}ari\'{c} \etal~\cite{vsaric2019single} use learnable deformations to help forecast future semantics from input RGB frames.
However, these methods do not explicitly consider dynamics of the scene.

While Jin~\etal~\cite{jin2017predicting} jointly predict flow and future semantics, some works explicitly warp deep features for future semantic segmentation~\cite{saric2020warp}.
Similarly, Terwilliger \etal~\cite{terwilliger2019recurrent} use an LSTM to estimate a flow field to warp the semantic output from an input frame.
However, by warping in output space -- rather than feature space -- their model is limited in its ability to reason about occlusions and depth.
While flow improves the modeling of the dynamic world, these methods only consider the dynamics at the pixel-level. Instead, we model dynamics at the object level. 

Recent methods ~\cite{qi20193d, vora2018future, xu2018structure, hoyer2019short} estimate future frames by reasoning about shape, egomotion, and foreground motion separately.
However, none of these methods reason explicitly about individual instances, while our method yields a full future panoptic segmentation forecast.

\noindent\textbf{Forecasting future instances:} 
Recent approaches for forecasting instance segmentation use a conv net to regress the deep features corresponding to the future instance segmentation \cite{luc2018predicting} or LSTMs \cite{sun2019predicting}.
Couprie \etal \cite{couprie2018joint} use a conv net to forecast future \emph{instance contours} together with an instance-wise semantic segmentation to estimate future instance segmentation. 
Their method only estimates foreground and not background semantics.
Several works have focused on anticipating future pose and location of specific object types, often people \cite{mangalam2020disentangling, graber2020dynamic}.
Ye \etal \cite{ye2019compositional} forecast future RGB frames by modeling each foreground object separately. 
Unlike these works, we anticipate both instance segmentation masks for foreground objects and background semantics for future time steps.

\section{Panoptic Segmentation Forecasting}

We introduce \emph{Panoptic Segmentation Forecasting}, a new task  which requires to anticipate the panoptic segmentation for a future, unobserved scene. Different from classical panoptic segmentation which \emph{analyzes} an observation, panoptic segmentation forecasting asks to \emph{anticipate} what the panoptic segmentation   looks like at a later time. 

Formally, given a series of $T$ RGB images $I_1, \dots, I_T$ of height $H$ and width $W$, the task is to anticipate the panoptic segmentation $S_{T+F}$ that corresponds to an unobserved future frame $I_{T+F}$ at a fixed number of timesteps $F$ from the last observation recorded at time $T$. Each pixel in $S_{T+F}$ is assigned a class $c \in \{1, \dots, C\}$ and an instance ID.

\begin{figure*}[t]
    \centering
    \includegraphics[width=\textwidth]{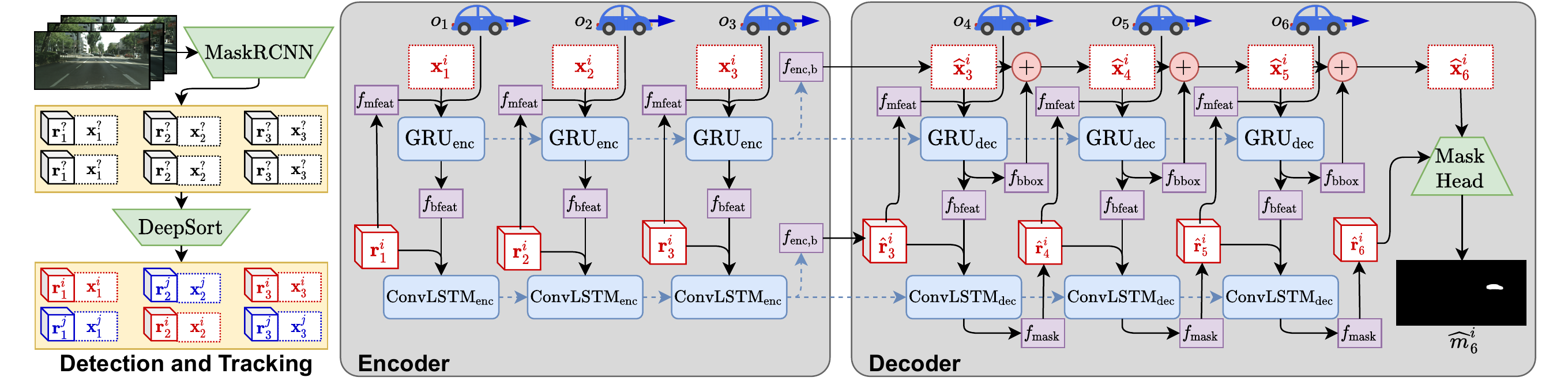}
    \vspace{-0.4cm}
    \caption{\textbf{The `Things' forecasting model.} This produces instance masks $\widehat{m}_{T+F}^i$ for each instance $i$ at target frame $T+F$. These masks are obtained from input images $I_1, \dots, I_T$ via the following procedure:  images are used to produce bounding box feature $\mathbf{x}_t^i$ and mask features $\mathbf{r}_t^i$ using MaskR-CNN and DeepSort (left). These features  are then input into an encoder to capture the instance motion history (middle). Encoder outputs are used to initialize a decoder, which predicts the features $\widehat{\mathbf{x}}_{T+F}^i$ and $\widehat{\mathbf{r}}_{T+F}^i$ for target frame $T+F$ (right). These features are passed through a mask prediction head to produce the final output $\widehat{m}_{T+F}^i$. Here, $T=3$ and $T+F = 6$.}
    \label{fig:foreground_model}
    \vspace{-0.4cm}
\end{figure*}

\subsection{Method}\label{sec:Method}

Anticipating the state of a future unobserved scene requires to understand the dynamics of its components. `Things' like cars, pedestrians, \etc  often traverse the world `on their own.' Meanwhile, stationary `stuff' changes  position in the image due to  movement of the observer camera. 
Because of this distinction, we expect the dynamics of `things' and `stuff' to differ. 
Therefore, we develop a model comprised of two components, one for the dynamics of detected `things'  and one for the rest of the `stuff.'

In addition to RGB images, we assume access to camera poses $o_1, \dots, o_T$ and depth maps $d_1, \dots, d_T$ for input frames. 
Camera poses can come from odometry sensors or estimates of off-the-shelf visual SLAM methods \cite{ORBSLAM3_2020}. 
We obtained our depth maps from input stereo pairs \cite{gu2020cascade} (these could also be estimated from single frames \cite{watson2019depthints}).

An overview of our panoptic segmentation forecasting  is shown in \figref{fig:overview}. The method consists of four stages:

\noindent{\bf 1) `Things' forecasting}  (\secref{sec:fg}): For each instance $i$, we extract foreground instance tracks $l^i$ from the observed input images  $I_1, \dots, I_T$. We use these tracks in our model to anticipate a segmentation mask $\widehat{m}^i_{T+F}$ and depth $\widehat{d}^i_{T+F}$  for the unobserved future frame at time $T+F$. 

\noindent{\bf 2) `Stuff' forecasting}  (\secref{sec:bg}): We predict the change in the background scene as a function of the anticipated camera motion, producing a background semantic output $\widehat{m}^B_{T+F}$ for the unobserved future frame $I_{T+F}$.

\noindent{\bf 3) Aggregation} (\secref{sec:agg}): We aggregate foreground `things' instance forecasts $\widehat{m}^i_{T+F}$ and background scene forecast $\widehat{m}^B_{T+F}$, producing the final panoptic segmentation output $S_{T+F}$ for future frame $I_{T+F}$.

\noindent{\bf 4) Odometry anticipation}  (\secref{sec:odom}): To better handle situations where we do not know future odometry, we train a model to forecast odometry from the input motion history.

\vspace{-0.4cm}
\subsubsection{`Things' forecasting:}
\label{sec:fg}

The foreground prediction model, sketched in \figref{fig:foreground_model}, first locates the instance locations $l^i$ within the input sequence. These tracks are each then independently processed by an encoder which captures their motion and appearance history. Encoder outputs are then used to initialize the decoder, which predicts the appearance and location of instances for future frames, including depth $\widehat{d}^i_{T+F}$. These are processed using a mask prediction model to produce the final instance mask $\widehat{m}^i_{T+F}$.

The output of the foreground prediction model is a set of estimated binary segmentation masks $\widehat{m}_{T+F}^i \in \{0, 1\}^{H\times W}$ representing the per-pixel location for every detected instance $i$ at frame $T+F$. Formally we obtain the  mask via
\begin{align}
    \tilde{m}^i_{T+F} &= \text{MaskOut}\left( \widehat{\mathbf{r}}^i_{T+F} \right), \\
    \widehat{m}^i_{T+F} &= \text{Round}\left(\text{Resize}\left(\tilde{m}^i_{T+F}, \widehat{\textbf{x}}_{T+F}^i\right) \right).
\end{align}
Here, in a first step, $\text{MaskOut}$ uses a small convolutional network (with the same architecture as the mask decoder of \cite{he2017mask}) to obtain  fixed-size segmentation mask probabilities $\tilde{m}^i_{T+F} \in [0, 1]^{28 \times 28}$ from  a mask feature tensor $\widehat{\mathbf{r}}^i_{T+F} \in \mathbb{R}^{256\times 14 \times 14}$. 
In a second step, $\text{Resize}$ scales this mask to the size of the predicted bounding box represented by the bounding box representation vector $\widehat{\textbf{x}}_{T+F}^i$ using bilinear interpolation while filling all remaining locations with $0$. The bounding box information vector $\widehat{\textbf{x}}_{T+F}^i \coloneqq [cx, cy, w, h, d, \Delta cx, \Delta cy, \Delta w, \Delta h, \Delta d]$ contains object center coordinates, width, and height, which are used in Resize, and also an estimate of the object's distance from the camera, and the changes of these quantities from the previous frame, which will be useful later. 
The output depth $\widehat{d}^i_{T+F}$ is also obtained from this vector.

\paragraph{Decoder.}
To anticipate the bounding box information vector $\widehat{\textbf{x}}_{T+F}^i$ and its appearance $\widehat{\mathbf{r}}^i_{T+F}$, we use a \textbf{decoder}, as shown on the right-hand-side of \figref{fig:foreground_model}. It is comprised primarily of two recurrent networks: a GRU~\cite{cho2014learning} which models future bounding boxes and a ConvLSTM \cite{xingjian2015convolutional} which models the future mask features. 
Intuitively, the GRU and ConvLSTM update hidden states $h_{b,t}^{i}$ and $h_{m,t}^{i}$, representing the current location and appearance of instance $i$, as a function of the bounding box features $\widehat{\mathbf{x}}_{t-1}$ and mask features $\widehat{\mathbf{r}}_{t-1}^i$ from the previous time step. These states are used to predict location and appearance features for the current timestep, which are then autoregressively fed into the model to forecast into the future; this process continues for $F$ steps until reaching the target time step $T+F$.
More formally,
\begin{align}
    h_{b,t}^{i} &= \text{GRU}_\text{dec}([\widehat{\mathbf{x}}_{t-1}, o_{t}, f_\text{mfeat}(\widehat{\mathbf{r}}_{t-1}^i)], h_{b,t-1}^{i}),\\
    \widehat{\mathbf{x}}_{t}^i &= \widehat{\mathbf{x}}_{t-1}^i + f_\text{bbox}(h_{b,t}^{i}), \\
    h_{m,t}^{i} &= \text{ConvLSTM}_\text{dec}([\widehat{\mathbf{r}}_{t-1}^i, f_{\text{bfeat}}(h_{b,t}^{i})], h_{m,t-1}^{i}), \\
    \widehat{\mathbf{r}}_{t}^i &= f_{\text{mask}}(h_{m,t}^{i}),
\end{align}
for $t \in \{T+1, \dots, T+F\}$, where $o_{t}$ represents the odometry at time $t$, $f_\text{bbox}$ and $f_\text{bfeat}$ are multilayer perceptrons, and $f_\text{mask}$ and $f_\text{mfeat}$ are $1\times1$ convolutional layers.

\paragraph{Encoder.}
The decoder uses bounding box hidden state $h^i_{b,T}$, appearance feature hidden state $h^i_{m,T}$, and estimates of the bounding box features $\widehat{\textbf{x}}^i_{T}$ and mask appearance features $\widehat{\textbf{r}}^i_{T}$ for the most recently observed frame $I_T$. We obtain these quantities from an \textbf{encoder} which processes the motion and appearance history of  instance $i$. Provided with bounding box features $\mathbf{x}_t^i$, mask features $\mathbf{r}_t^i$, and odometry $o_t$ for input time steps $t\in \{1, \dots, T\}$, the encoder computes the aforementioned quantities via 
\begin{align}
    h_{b,t}^{i} &= \text{GRU}_\text{enc}([\mathbf{x}^i_{t-1}, o_{t-1}, f_\text{mfeat}(\mathbf{r}_{t-1}^i)], h_{b,t-1}^{i}),\\
    h_{m,t}^{i} &= \text{ConvLSTM}_\text{enc}([\mathbf{r}_{t-1}^i, f_{\text{bfeat}}(h_{b,t}^{i})], h_{m,t-1}^{i}).
\end{align}
Intuitively, the bounding box encoder is a GRU which processes input bounding box features $\mathbf{x}^i_{t}$, odometry $o_{t}$, and a transformation of mask features $\mathbf{r}^i_t$ to produce box state representation $h^i_{b,T}$. Additionally, the mask appearance encoder is a ConvLSTM which processes input mask features $\mathbf{r}^i_t$ and the representation of the input bounding box features $h^i_{b,t}$ produced by the bounding box encoder to obtain mask state representation $h^i_{m,T}$.

The estimated mask and bounding box feature estimates for the final input time step $T$ are  computed by processing the final encoder hidden states via
\begin{align}
    \widehat{\mathbf{x}}_{T}^i &= f_\text{enc,b}(h_{b,T}^{i}), \text{~~and~~}
    \widehat{\mathbf{r}}_{T}^i = f_{\text{enc,m}}(h_{m,T}^{i}),
    \label{eq:pred}
\end{align} 
where $f_\text{enc,b}$ is a multilayer perceptron and $f_{\text{enc,m}}$ is a $1 \times 1$ convolution. These estimates are necessary because occlusions can prevent access to  location and appearance for time step $T$ for some object instances. In those cases, use of \equref{eq:pred} is able to fill the void.

\paragraph{Tracking.}
The encoder operates on estimated instance tracks/locations $l^i \coloneqq \{c^i, (\mathbf{x}_{t}^i, \mathbf{r}_{t}^i)|_{t=1}^T\}$ which consist of object class $c^i$, bounding box features $\mathbf{x}_{t}^i$ and mask features $\mathbf{r}_t^i$ for all instances in the input video sequence $I_1, \dots, I_T$. Obtaining these involves two steps: 1) we run MaskR-CNN~\cite{he2017mask} on every input frame to find the instances; 2) we link instances across time using 
DeepSort~\cite{wojke2017simple}. For a given tracked instance $i$, we use outputs provided by MaskR-CNN, including predicted class $c^i$, bounding boxes $\mathbf{x}_{t}^i$, and mask features $\mathbf{r}_{t}^i$ extracted after the ROIAlign stage. The distance $d$ within $\mathbf{x}^i_t$ refers to the median value of the input depth map $d_t$ at locations corresponding to the estimated instance segmentation mask found by MaskR-CNN for instance $i$ in input frame $t$. A given object instance may not be found in all input frames, either due to the presence of occlusions or because it has entered or left the scene. In these cases, we set the inputs to  an all zeros tensor. 

\begin{figure*}[t]
    \centering
    \footnotesize
    \setlength\tabcolsep{0.5pt}
    \renewcommand{\arraystretch}{0.5}
    \begin{tabular}{ccccc}
        Last Seen Image &  Oracle & Flow & Hybrid & Ours\\
         \includegraphics[width=0.19\textwidth]{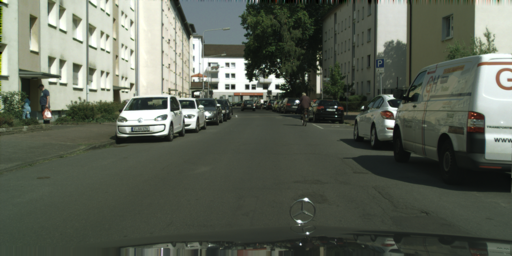}&
         \includegraphics[width=0.19\textwidth]{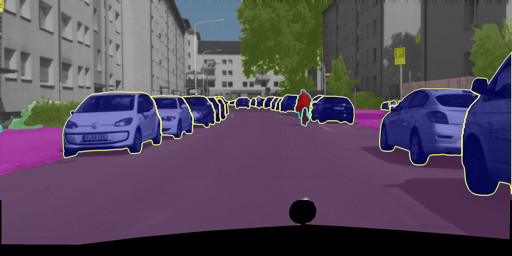}& \includegraphics[width=0.19\textwidth]{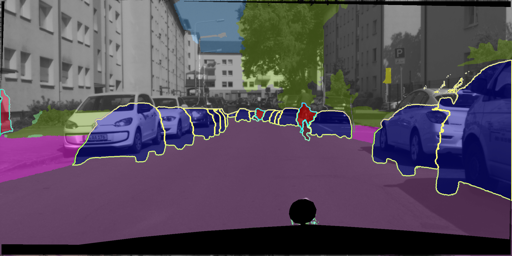}& \includegraphics[width=0.19\textwidth]{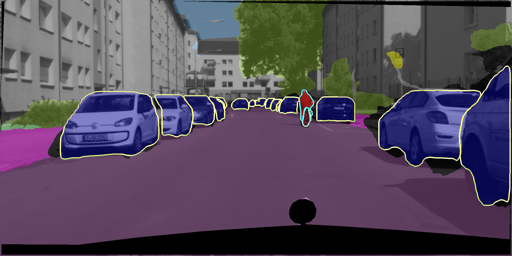}& \includegraphics[width=0.19\textwidth]{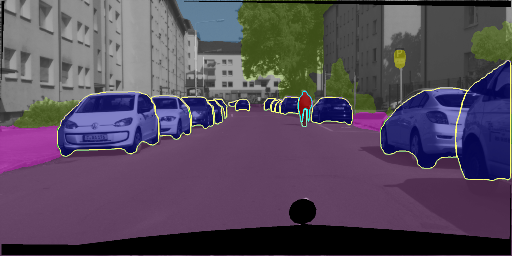}\\
         \includegraphics[width=0.19\textwidth]{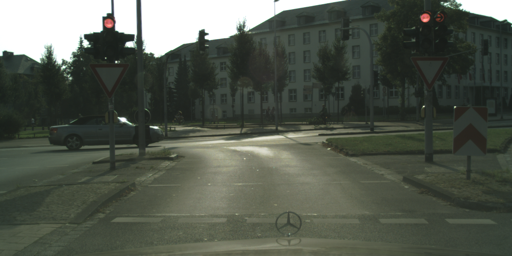}&
         \includegraphics[width=0.19\textwidth]{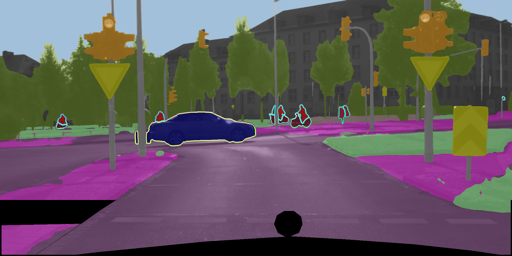}& \includegraphics[width=0.19\textwidth]{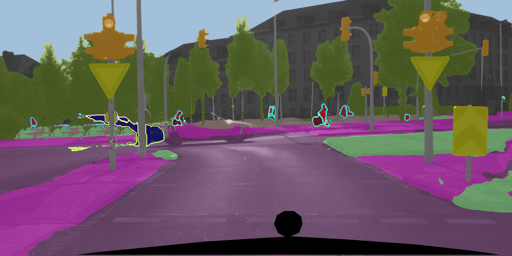}& \includegraphics[width=0.19\textwidth]{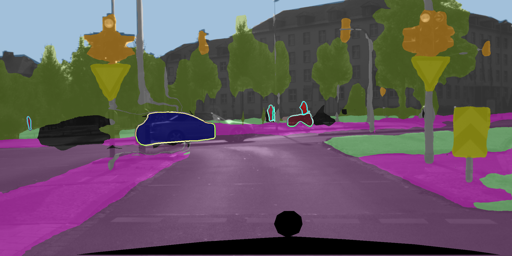}& \includegraphics[width=0.19\textwidth]{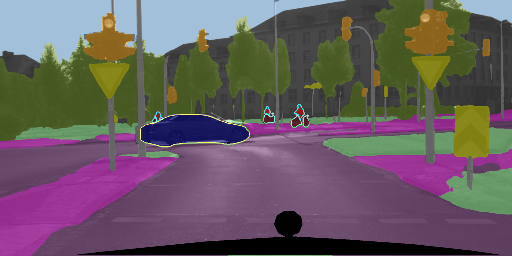}\\
          \includegraphics[width=0.19\textwidth]{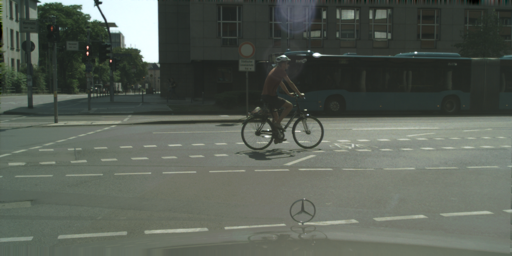}&
         \includegraphics[width=0.19\textwidth]{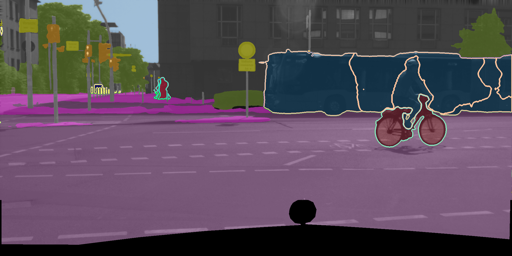}& \includegraphics[width=0.19\textwidth]{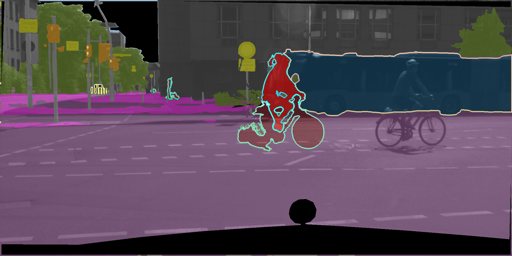}& \includegraphics[width=0.19\textwidth]{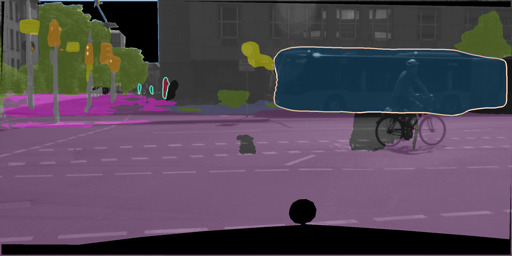}& \includegraphics[width=0.19\textwidth]{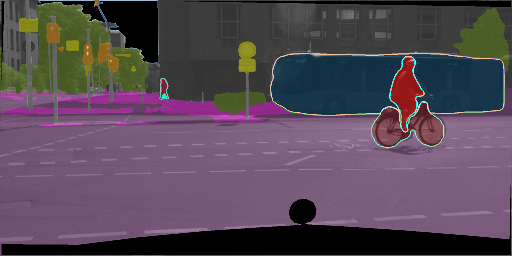}\\
         \includegraphics[width=0.19\textwidth]{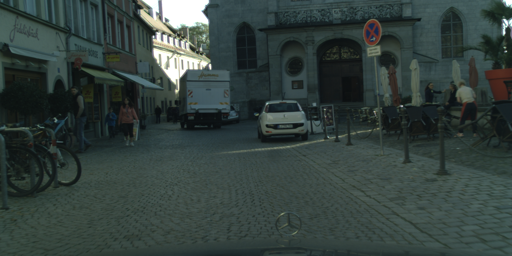}&
         \includegraphics[width=0.19\textwidth]{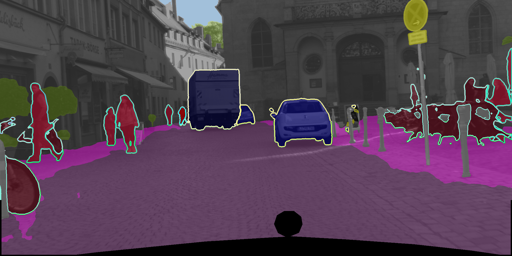}& \includegraphics[width=0.19\textwidth]{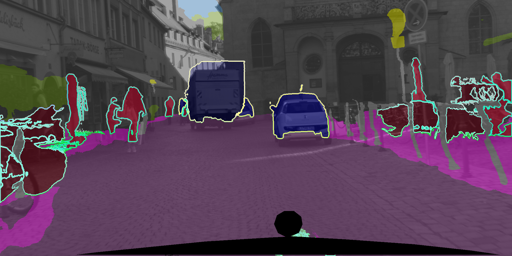}& \includegraphics[width=0.19\textwidth]{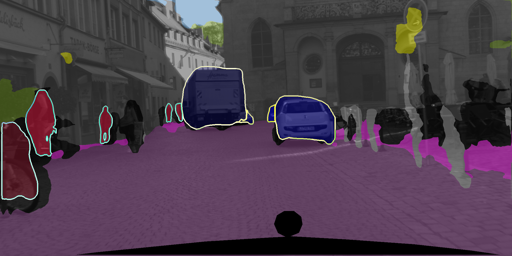}& \includegraphics[width=0.19\textwidth]{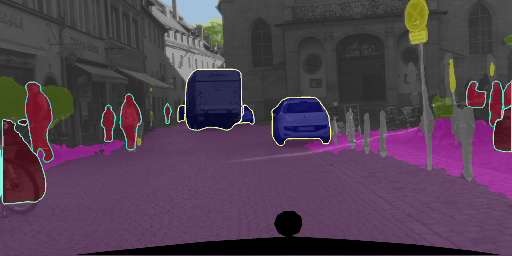}\\
    \end{tabular}
    \caption{\textbf{Mid-term panoptic segmentation forecasts on Cityscapes.} Compared to \emph{Hybrid}, our approach produces  more well-defined silhouettes for instance classes (see the cars in the 1st row or the pedestrians in the 4th row), and handles instances with large motion much better \emph{Hybrid} -- the car in the 2nd row is not predicted to have moved sufficiently; the cyclist in the 3rd row is not predicted at all. Since \emph{Flow} does not model instance-level trajectory, the `things' are no longer intact in the forecasts.}
    \label{fig:panoptic_viz}
    \vspace{-0.4cm}
\end{figure*}

Note that it is possible for instances to be missed during the detection phase. We largely observe this to happen for static objects such as groups of bicycles parked on a sidewalk (for instance, the right side of our prediction in the fourth row of \figref{fig:panoptic_viz}). 
One solution is to consider these instances as part of the background forecasting. However, in our experiments, we found that treating all the missed instances as background degraded our performance because some instances are actually dynamic. Thus, in this paper, we choose not to recover these instances.

\noindent{\textbf{Losses.}} To train the foreground model, we provide input location and appearance features, predict their future states, and regress against their pseudo-ground-truth future states. More specifically, the losses are computed using the estimated bounding boxes $\mathbf{x}_t^i$ and instance features $\mathbf{r}_t^i$  found by running instance detection and tracking on  future frames. Note that losses are computed on intermediate predictions as well, which permits to properly model motion and appearance of instances across all future time steps.
Our foreground model loss is a weighted sum of mean squared error and L1 losses. 
See appendix \secref{sec:sup_foreground_loss} for full details.

\vspace{-0.4cm}
\subsubsection{`Stuff' forecasting:}
\label{sec:bg}

The background `stuff' forecasting is tasked with predicting a semantic output $\widehat{m}^B_{T+F} \in \{1, \dots, C_\text{stuff}\}^{H\times W}$ for every pixel in the target frame $T+F$. 
We assume they correspond to the static part of the scene, \ie, background changes in the images are caused solely by camera motion.

We predict the background changes by back-projecting 3D points from the background pixels in frame $t$ given depth $d_t$ and camera intrinsics, transforming with ego-motion $o_t$, and projecting to frame $T+F$.
This process establishes pixel correspondences between input frame $t$ and target frame $T+F$. After running a pre-trained semantic segmentation model on frame $I_t$ to get semantic segmentation $m_t$, we use these correspondences to map the semantic labels from $m_t$, which correspond to ``stuff'' classes, to pixels in frame $T+F$ and maintain their projected depth at this frame.
We denote the projected semantic map as $\tilde{m}^B_t$ and the projected depth as $\tilde{d}^B_t$.
However, due to 1) sparsity of the point clouds, and 2) lack of information in regions which were previously occluded by foreground objects or were not previously in-frame, only a subset of pixels in $\tilde{m}^B_t$ are assigned a label. 
Therefore, we apply a refinement model that takes in $(\tilde{m}^B_t, \tilde{d}^B_t)$ from all input frames to complete the semantic segmentation map $\widehat{m}^B_{T+F}$.

\noindent{\textbf{Losses.}} To train the background refinement model, we use a cross-entropy loss applied at pixels which do not correspond to foreground objects in the target frame.
This encourages the output of the refinement network to match the ground truth semantic segmentation at each pixel.
We formalize this in appendix \secref{sec:sup_background_loss}.

\begin{table*}[t]
    \vspace{-0.2cm}
    \footnotesize
      \centering
      \resizebox{1.0\textwidth}{!}{
      \begin{tabular}{l ccc ccc ccc c ccc ccc ccc}
    \toprule
    & \multicolumn{9}{c}{Short term: $\Delta t = 3$} && \multicolumn{9}{c}{Mid term: $\Delta t = 9$} \\
    & \multicolumn{3}{c}{All} & \multicolumn{3}{c}{Things} & \multicolumn{3}{c}{Stuff}
    && \multicolumn{3}{c}{All} & \multicolumn{3}{c}{Things} & \multicolumn{3}{c}{Stuff} \\
    & PQ & SQ & RQ & PQ & SQ & RQ & PQ & SQ & RQ && PQ & SQ & RQ & PQ & SQ & RQ & PQ & SQ & RQ\\
    \midrule
    Panoptic Deeplab (Oracle)$\dagger$ & $60.3$ & $81.5$ & $72.9$    & $51.1$ & $80.5$ & $63.5$    & $67.0$ & $82.3$ & $79.7$         && $60.3$ & $81.5$ & $72.9$ &$51.1$ & $80.5$ & $63.5$    & $67.0$ & $82.3$ & $79.7$   \\
    \midrule
   Panoptic Deeplab (Last seen frame)  & $32.7$ & $71.3$ & $42.7$      & $22.1$ & $68.4$ & $30.8$ & $40.4$  & $73.3$ & $51.4$  && $22.4$ & $68.5$ & $30.4$ & $10.7$ & $65.1$ & $16.0$ & $31.0$ & $71.0$ & $40.9$ \\
    Flow & $41.4$ & $73.4$ & $53.4$ & $30.6$ & $70.6$ & $42.0$ & $49.3$ & $75.4$ & $61.8$ && $25.9$ & $69.5$ & $34.6$ & $13.4$ & $67.1$ & $19.3$ & $35.0$ & $71.3$ & $45.7$\\
    Hybrid \cite{terwilliger2019recurrent} (bg) and \cite{luc2018predicting} (fg)  &  $43.2$ & $74.1$ & $55.1$ & $35.9$ & $72.4$ & $48.3$ & $48.5$ & $75.3$ & $60.1$ & & $29.7$ & $69.1$ & $39.4$ & $19.7$ &  $66.8$ &  $28.0$ & $37.0$ &  $70.8$ &  $47.7$\\
    \textbf{Ours}  & $\textbf{49.0}$ & $\textbf{74.9}$ & $\textbf{63.3}$ & $\textbf{40.1}$ & $\textbf{72.5}$ & $\textbf{54.6}$ & $\textbf{55.5}$ & $\textbf{76.7}$ & $\textbf{69.5}$       & & $\textbf{36.3}$ & $\textbf{71.3}$ & $\textbf{47.8}$ & $\textbf{25.9}$ & $\textbf{69.0}$ & $\textbf{36.2}$ & $\textbf{43.9}$ & $\textbf{72.9}$ & $\textbf{56.2}$\\
    \bottomrule
      \end{tabular}}
      \vspace{-0.1cm}
      \caption{\textbf{Panoptic segmentation forecasting evaluated on the Cityscapes validation set}. 
      $\dagger$ has access to the RGB frame at $\Delta t$. Higher is better for all metrics.}
      \label{tab:panoptic}
      \vspace{-0.1cm}
 \end{table*}

\vspace{-0.2cm}
\subsubsection{Aggregation:}
\label{sec:agg}

This step combines foreground instance segmentations $\widehat{m}_{T+F}^i$, classes $c^i$,  depths $d^i_{T+F}$ and background semantic prediction $\widehat{m}_{T+F}^B$ into the final future panoptic segmentation $S_{T+F}$. 
For simplicity, we assume that all foreground objects are located in front of all background components. We found this to be valid in most cases. Thus, to combine the foreground and background, we `paste' foreground instances  in order of decreasing predicted instance depth on top of the background. This approach is presented visually in \figref{fig:overview}, right, and described in more detail by \algref{alg:aggregation} in the appendix. 

\vspace{-0.2cm}
\subsubsection{Egomotion estimation:}
\label{sec:odom}

A large contributor to the observed motion of a scene is  the movement of the recording camera. Properly modeling this movement is critical for accurate results. 
Here, we consider two scenarios: 1) an `active' scenario where the model has access to the planned motion of an autonomous agent; 2) a `passive' scenario in which the camera is controlled by an external agent and hence the model is not provided with the future motion. 

In the active scenario, we use  the speed and yaw rate of the camera from the dataset, which we process into the forms required by the foreground and background models. See appendix \secref{app:odometry}  for more details.

In the passive scenario, we use a GRU to predict the future camera motion as a function of its past movement. More formally and as sketched in \figref{fig:overview}(left),
\begin{align}
    h_{o,t+1} \!=\! \text{GRU}_\text{cam}\left(\widehat{o}_{t}, h_{o,t}\right)  
    \text{~and~} \widehat{o}_{t+1} \!=\! f_{\text{cam}}(h_{o,t+1}),
\end{align}
where $f_{\text{cam}}$ is a multilayer perceptron. For input time steps, \ie, $t\in\{1,\dots, T\}$, we use known camera motion $o_t$ as model input. For future time steps, \ie, $t\in\{T+1, \dots, T+F\}$, we use predicted camera motion $\widehat{o}_t$ as input.

\section{Evaluation}

We establish the first  results for the task of panoptic segmentation forecasting by comparing our developed approach to several baselines. We also provide ablations to demonstrate the importance of our modeling decisions. We additionally evaluate on the tasks of semantic segmentation forecasting and instance segmentation forecasting to compare our method to prior work on established tasks. 

\noindent\textbf{Data:} 
To  evaluate panoptic segmentation forecasting, we need a dataset which contains both semantic and instance information as well as  entire video sequences that lead up to the annotated frames.  Cityscapes~\cite{cordts2016cityscapes}  fulfills these requirements and has been used in prior work for semantic and instance forecasting. This dataset consists of 5000 sequences of 30 frames each,  spanning approximately 1.8 seconds. The data were recorded from a car driving in urban scenarios, and semantic and instance annotations are provided for the 20th frame in each sequence. Following standard practice for prior work in forecasting segmentations \cite{luc2017predicting, luc2018predicting, terwilliger2019recurrent, saric2020warp}, all experiments presented here are run on the validation data; a limited set of evaluations on test data are presented in appendix~\secref{sec:appx_eval}. 

To match prior work \cite{luc2017predicting, luc2018predicting, saric2020warp}, we use every third frame as input and evaluate  two different scenarios: short-term forecasting looks 3 frames (${\sim}0.18$s)  and medium-term forecasting looks 9 frames (${\sim}0.53$s) into the future.  All metrics are computed on the 20th frame of the sequence. We use an input length of $T=3$. We hence use frames $11$, $14$, and $17$ as input for short-term experiments and frames $5$, $8$, and $11$ as input for medium-term experiments.

\subsection{Panoptic Segmentation Forecasting}
\label{sec:exp_panoptic}
\noindent{\textbf{Metrics.}} We compare all approaches using  metrics introduced in prior work~\cite{kirillov2019panoptic} on  panoptic segmentation. These metrics require to first compute matches between predicted segments and ground truth segments. A match between a predicted segment and a ground truth segment of the same class is a true positive if their intersection over union (IoU) is larger than $0.5$. Using these matches, three  metrics are considered: \emph{segmentation quality} (SQ), which is the average IoU of true positive matched segments, \emph{recognition quality} (RQ), which is the F1 score computed over matches, and \emph{panoptic quality} (PQ), which is the product of SQ and RQ. All of these metrics are computed per class and then averaged to compute the final score.

\noindent{\textbf{Baselines.}}
To compare  our approach against baselines on the novel task of panoptic segmentation forecasting, we use: \\
\emph{Panoptic Deeplab (Oracle)}: we apply the Panoptic Deeplab model~\cite{Cheng2020panoptic-deeplab} to \emph{analyze} the target frame. This represents an upper bound on performance, as it has direct access to future information. \\
\emph{Panoptic Deeplab (Last seen frame)}: we apply the same Panoptic Deeplab model to the most recently observed frame. This represents a model that assumes no camera or instance motion. \\
\emph{Flow}: Warp the panoptic segmentation analyzed at the last observed frame using optical flow~\cite{ilg2017flownet} computed from the last two observed frames. \\
\emph{Hybrid Semantic/Instance Forecasting}: We fuse predictions made by a semantic segmentation forecasting model~\cite{terwilliger2019recurrent} and an instance segmentation forecasting model~\cite{luc2018predicting} to create a panoptic segmentation for the target frame.

\begin{table}
\setlength{\tabcolsep}{3pt}
    \footnotesize
      \centering
      \begin{tabular}{lcccccc}
    \toprule
    & \multicolumn{3}{c}{$\Delta t = 3$} & \multicolumn{3}{c}{$\Delta t = 9$} \\
    & PQ & SQ & RQ & PQ & SQ & RQ \\
    \midrule
    \textbf{Ours} & $\textbf{49.0}$ & $74.9$ & $\textbf{63.3}$ & $\textbf{36.3}$ & $71.3$ & $\textbf{47.8}$\\
    1) w/ Hybrid bg \cite{terwilliger2019recurrent} & $45.0$ & $74.1$ & $57.9$ & $32.4$ & $70.1$ & $42.9$ \\
    2) w/ Hybrid fg \cite{luc2018predicting} & $47.3$ & $74.8$ & $60.7$ & $33.4$ & $70.4$ & $43.9$ \\
    3) w/ linear instance motion & $40.2$ & $73.7$ & $52.1$ & $27.9$ & $70.1$ & $36.6$ \\
    4) fg w/o odometry & $48.8$ & $75.1$ & $62.8$ & $35.3$ & $71.1$ & $46.5$ \\
    5)  w/ ORB-SLAM odometry & $48.6$ & $75.0$ & $62.5$ & $36.1$ & $71.3$ & $47.5$ \\
    6)  w/ SGM depth  & $48.8$ & $\textbf{75.2}$ & $62.8$ & $36.1$ & $\textbf{71.4}$ & $47.3$ \\
    7)  w/ monocular depth & $47.5$ & $74.8$ & $61.0$ & $34.8$ & $70.9$ & $45.8$ \\ 
    \midrule
    w/ ground truth future odometry  &  $49.4$ & $75.2$ & $63.5$ & $39.4$ & $72.1$ & $51.6$ \\
    \bottomrule
      \end{tabular}
      \vspace{-0.1cm}
      \caption{\textbf{Validating our design choices} using Cityscapes. Higher is better for all metrics. All approaches use predicted future odometry unless otherwise specified.}
      \label{tab:ablations}
      \vspace{-0.5cm}
 \end{table}

\noindent{\textbf{Results.}} 
The results for all models on the panoptic segmentation forecasting task are presented in \tabref{tab:panoptic}. We outperform all non-oracle approaches on the PQ, SQ, and RQ metrics for both short-term and mid-term settings. The improvements to PQ and RQ show that our model better captures the motion of all scene components, including static background `stuff' regions and dynamic `things.' In addition, the improvements to SQ
imply that the per-pixel quality of true positive matches are not degraded. 
The \emph{Flow} model performs worse than either \emph{Hybrid} or our approach, which demonstrates that a simple linear extrapolation of per-pixel input motion is not sufficient to capture the scene and object movement.
The fact that the gap between ours and \emph{Hybrid} on `things' PQ grows between the short- and mid-term settings shows the strength of our foreground model (\secref{sec:fg}) on anticipating object motion at longer time spans.
\figref{fig:panoptic_viz} compares results to baselines. Our approach produces better defined object silhouettes and handles large motion better than the baselines. 

\noindent{\textbf{Ablations.}} \tabref{tab:ablations} shows results for ablation experiments which analyze the impact of our modeling choices: 1)~\emph{w/Hybrid bg} uses our foreground model, but replaces our background model with the one from~\cite{terwilliger2019recurrent}; 2)~\emph{w/Hybrid fg} uses our background model, but replaces our foreground model with the one  from~\cite{luc2018predicting}; 3)~\emph{w/linear instance motion} replaces the foreground forecasting model with a simple model assuming linear instance motion and no mask appearance change; 
4)~\emph{fg w/o odometry} does not use odometry as input to the foreground model; 5)~\emph{w/ ORB-SLAM odometry} uses input odometry obtained from~\cite{ORBSLAM3_2020}; 6)~\emph{w/ SGM depth} uses depths obtained from SGM~\cite{hirschmuller2005accurate} provided by~\cite{cordts2016cityscapes} as input to the model; and 7)~\emph{w/ monocular depth} uses a monocular depth prediction model~\cite{godard2019digging}, finetuned on Cityscapes, to obtain input depth.
Ablations 1) and 2) show that our improved model performance is due to the strength of both our foreground and background components.
Ablation 3) shows that joint modeling of instance motion and appearance mask is  key to success.
4) shows that odometry inputs help the model predict foreground locations better, and 5) demonstrates our method works well with odometry computed directly from input images.
6) and 7) suggest that our approach benefits from more accurate depth prediction, but it also works well with depth inputs obtained using single-frame methods.

\subsection{Semantic Segmentation Forecasting}
\begin{table}[t]

\footnotesize
  \centering
  \begin{tabular}{lcccc}
\toprule
& \multicolumn{2}{c}{Short term: $\Delta t = 3$} & \multicolumn{2}{c}{Mid term: $\Delta t = 9$} \\
Accuracy (mIoU) & All & MO & All & MO \\
\midrule
Oracle                 &  $80.6$ & $81.7$ & $80.6$ & $81.7$ \\
\midrule
Copy last   & $59.1$  &  $55.0$ & $42.4$ & $33.4$ \\
3Dconv-F2F \cite{chiu2020segmenting}          &  $57.0$ & / & $40.8$ & /  \\
Dil10-S2S \cite{luc2017predicting}          &  $59.4$ & $55.3$ & $47.8$ & $40.8$ \\
LSTM S2S \cite{rochan2018future}           &  $60.1$ & / & / & / \\
Bayesian S2S \cite{bhattacharyya2019bayesian}        &  $65.1$ & / & $51.2$ & / \\
DeformF2F \cite{vsaric2019single}          &  $65.5$ & $63.8$ & $53.6$ & $49.9$ \\
LSTM M2M \cite{terwilliger2019recurrent}           &  $67.1$ & $65.1$ & $51.5$ & $46.3$ \\
F2MF \cite{saric2020warp}     &  $\textbf{69.6}$ & $\textbf{67.7}$ & $57.9$ & $\textbf{54.6}$ \\
\textbf{Ours}           &  $67.6$ & $60.8$ & $\textbf{58.1}$ & $52.1$\\
\bottomrule
  \end{tabular}
  \vspace{1pt}
  \caption{\textbf{Semantic forecasting results on the Cityscapes validation dataset}.  Baseline numbers, besides oracle and copy last, are from \cite{saric2020warp}. Higher is better for all metrics. Our model exploits stereo and odometry, which are provided by typical autonomous vehicle setups and are included in Cityscapes.}
  \label{tab:semantic}
  \vspace{-0.5cm}
 \end{table}
 
For a comprehensive comparison, we also assess our approach on the task of semantic segmentation forecasting. This task asks to anticipate the correct semantic class per pixel for the target frame. Unlike the panoptic segmentation evaluation, this task doesn't care about instances, \ie, good performance only depends on the ability to anticipate the correct semantic class for each pixel. We obtain semantic segmentation outputs from our model by discarding instance information and only retaining the semantics.

\noindent{\textbf{Metrics.}} Future semantic segmentation is evaluated using  intersection over union (IoU) of predictions compared to the ground truth, which are computed per class and averaged over classes. We additionally present an IoU score which is computed by averaging over `things' classes only (MO).

\noindent{\textbf{Baselines.}} We compare to a number of recent works which forecast semantic segmentations. Many of these approaches  anticipate the features of a future scene \cite{luc2017predicting, rochan2018future, bhattacharyya2019bayesian, vsaric2019single, chiu2020segmenting, saric2020warp}. 
LSTM M2M~\cite{terwilliger2019recurrent} anticipates the optical flow between the most recent frame and the target with a warping function transforming input semantics.  Different from these, we decompose the prediction into feature predictions for each individual instance as well as a transformation of background semantics before combining.
Additionally, these approaches do not use depth inputs and all except Bayesian S2S~\cite{bhattacharyya2019bayesian} do not use egomotion as input.

\noindent{\textbf{Results.}} The results for this task are given in \tabref{tab:semantic}. 
We outperform most models on standard IoU as well as MO IoU. Unlike all other baselines, our model is able to produce instance-level predictions for moving object classes, which is a more challenging objective. 

\vspace{-0.1cm}
\subsection{Instance Segmentation Forecasting}
\begin{table}[t]
\footnotesize
\centering
\begin{tabular}{lcccc}
\toprule
&\multicolumn{2}{c}{Short term: $\Delta t = 3$} & \multicolumn{2}{c}{Mid term: $\Delta t = 9$} \\
& AP & AP50 & AP & AP50 \\
\midrule
Oracle & $34.6$ & $57.4$ & $34.6$ & $57.4$\\
Last seen frame & $8.9$ & $21.3$ & $1.7$ & $6.6$\\
\midrule
F2F \cite{luc2018predicting} & $\textbf{19.4}$ & $\textbf{39.9}$ & $7.7$ & $19.4$ \\
Ours & $17.8$ & $38.4$ & $\textbf{10.0}$ & $\textbf{22.3}$\\
\bottomrule
\end{tabular}
\caption{\textbf{Instance segmentation forecasting on the Cityscapes validation dataset.} Higher is better for all metrics.}
\label{tab:inst_val}
\vspace{-0.25cm}
\end{table}

We also evaluate on instance segmentation forecasting, which only focuses on the `things' classes within Cityscapes. Future instance segmentation can be obtained from our model by disregarding all pixels corresponding to `stuff' classes from the panoptic forecasting output. 

\noindent{\textbf{Metrics.}} Instance segmentation is evaluated using two metrics~\cite{cordts2016cityscapes}: 1) Average Precision (AP) first averages over a number of overlapping thresholds required for matches to count as true positives and is then averaged across classes; 2) AP50 is the average precision computed using an overlap threshold of $0.5$ which is then averaged across classes. 

\noindent{\textbf{Baselines.}} There is very little prior work on instance segmentation forecasting. We compare to Luc~\etal~\cite{luc2018predicting}, who train a model to predict the features of the entire future scene using a convolutional model and obtain final instances by running these predicted features through the prediction heads of MaskR-CNN. 
Instead, our approach predicts an individual set of features for each instance found in the scene.
 
\noindent{\textbf{Results.}} \tabref{tab:inst_val} presents the results. 
We outperform prior work in the mid-term setting.
This indicates that modeling trajectory of individual instances has a higher potential on forecasting tasks. 
Since we use the same model created by Luc~\etal~\cite{luc2018predicting} as the `foreground' component of the \emph{Hybrid} baseline (\secref{sec:exp_panoptic}), \figref{fig:panoptic_viz} shows visual comparisons between these approaches. Again, our method gives higher-detailed instance contours and models objects with larger motion more accurately. Moreover, in some cases, F2F ``deletes'' some instances from the scene  (such as the cyclist in row $3$).

\subsection{Introspection} 
\begin{figure}[t]
\includegraphics[width=0.49\columnwidth]{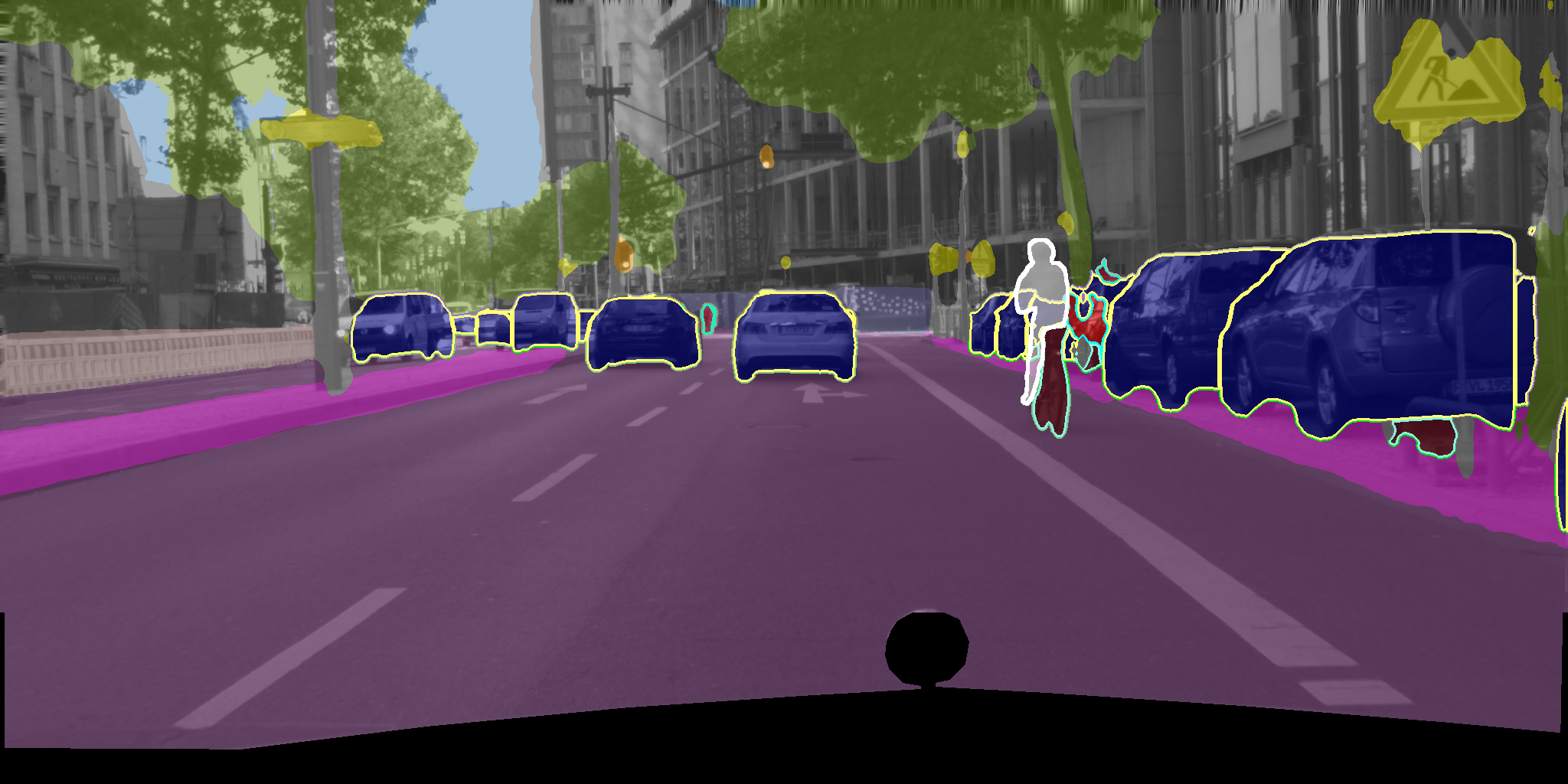}~
\includegraphics[width=0.49\columnwidth]{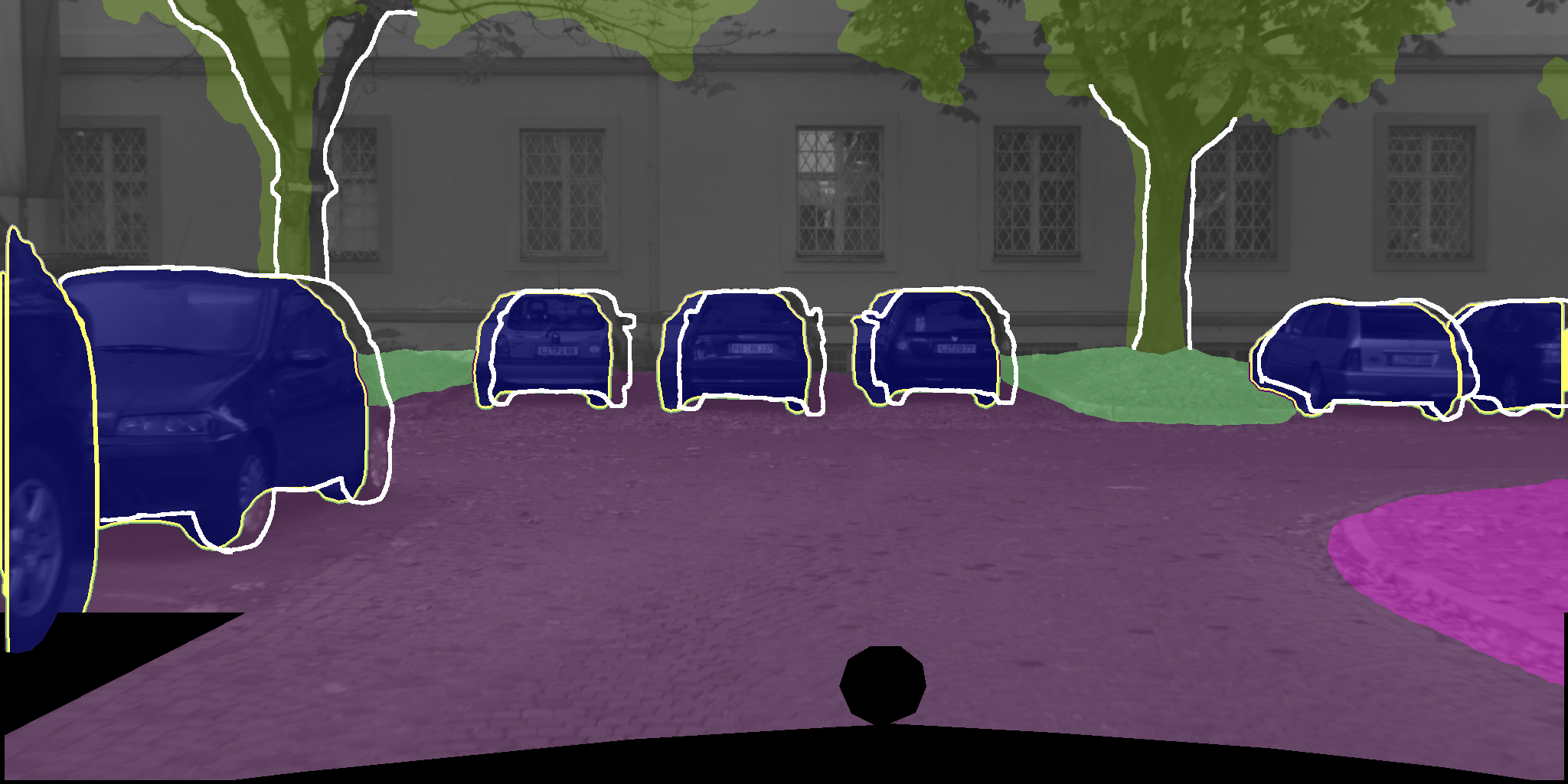}
\vspace{-0.5cm}
\caption{\textbf{Failure cases.} Left: the cyclist highlighted in white was missed by instance detection. Right: mispredicted odometry leads to misalignment between the forecast and the target image (the outlines of objects in the target image are shown in white).}
\label{fig:failures}
\vspace{-0.3cm}
\end{figure}

Why does our approach anticipate higher-fidelity instances than prior approaches? Many of these works attempt to predict future scenes by anticipating what a fixed-size feature tensor for the entire image will look like -- this is true for both semantic segmentation forecasting \cite{luc2017predicting, chiu2020segmenting, saric2020warp} and instance segmentation forecasting \cite{luc2018predicting}. Note, this  conflates camera motion, which objects are present in a scene, how these objects move, and how the appearance of objects and background components change as a function of the scene motion. This increases the complexity of the prediction. Instead, our method decomposes these components into individual parts: the foreground model anticipates how each object moves and how its appearance changes as a function of this motion; the background model captures how static scene components appear when the camera moves; and the odometry model anticipates likely future motion based on  past input.  Modeling each of these separately simplifies individual prediction. Additionally, we predict separate features for every individual instance, so its size scales with the number of instances present in a scene, while past approaches~\cite{luc2018predicting} use a fixed size representation regardless of the complexity of the scene.

The performance of our approach is hampered in some cases by failures in instance detection and tracking (examples in \figref{fig:failures}). At the moment, our model cannot properly recover from situations where the input is noisy. 
That being said, our approach immediately benefits from improvements in the areas of instance detection and tracking, which are very active fields of research \cite{bergmann2019tracking, bertasius2020classifying}.

\vspace{-0.2cm}
\section{Conclusions}
\vspace{-0.2cm}
We introduced the novel task `panoptic segmentation forecasting.' It requires to anticipate a per-pixel instance-level segmentation of `stuff' and `things'  for an unobserved future frame given as input a sequence of past frames. To solve this  task, we developed a model which anticipates trajectory and appearance of `things' and by reprojecting input semantics for `stuff.' We demonstrated that the method outperforms compelling baselines on panoptic, 
semantic and instance segmentation forecasting. 

\noindent\textbf{Acknowledgements:} This work is supported in part by NSF under Grant \#1718221, 2008387, 2045586, MRI \#1725729, and NIFA award 2020-67021-32799.

\clearpage

\bibliographystyle{ieee}
{\small
\bibliography{main}
}

\clearpage

\appendix
\twocolumn[
\noindent\textbf{\Large Appendix: Panoptic Segmentation Forecasting
}
\vspace{0.2cm}]

In this appendix, we first provide additional details on the classes contained within Cityscapes (\secref{app:breakdown}). This is followed by details deferred from the main paper for space reasons, including a description of how odometry is used by each component (\secref{app:odometry}), formal descriptions of the background modeling approach (\secref{app:background}), the aggregation method (\secref{app:aggregation}), used losses (\secref{app:losses}), and miscellaneous implementation details for all model components and baselines (\secref{app:impdetails}). This is followed by additional experimental results (\secref{sec:appx_eval}), including per-class panoptic segmentation metrics as well as metrics computed on test data for the tasks of panoptic, semantic, and instance segmentation forecasting. Finally, we present additional visualizations of model predictions (\secref{sec:appx_viz}).

Included in the supplementary material are videos visualizing results and a few of the baselines. These are described in more detail in \secref{sec:appx_viz}.

\section{`Things' and `Stuff' class breakdown in Cityscapes}
\label{app:breakdown}
The data within Cityscapes are labeled using $19$ semantic classes. The `things' classes are those that refer to individual objects which are potentially moving in the world and consist of person, rider, car, truck, bus, train, motorcycle, and bicycle. The remaining $11$ classes are the `stuff' classes and consist of road, sidewalk, building, wall, fence, pole, traffic light, traffic sign, vegetation, terrain, and sky.

\section{Odometry}
\label{app:odometry}
Cityscapes provides vehicle odometry $o_t$ at each frame $t$ as $[v_t, \Delta\theta_t]$, where $v_t$ is the speed and $\Delta \theta_t$ is the yaw rate. 
This odometry assumes the vehicle is moving on a flat ground plane with no altitude changes. For background forecasting (\secref{sec:bg}), a full 6-dof transform $H_t$ from frame $t$ and the target frame $T+F$ is required for projecting the 3d semantic point cloud $(\tilde{m}^B_t, \tilde{d}^B_t)$. 

To derive $H_t$, we use the odometry readings $o_{1, \dots, T+F}$ from frame $t$ to target frame $T+F$, the time difference between consecutive frames $\Delta t_{1, \dots, T+F}$, and the camera extrinsics $H^\text{cam}_\text{veh}$ (\ie, transformation from the vehicle's coordinate system to the camera coordinate system)  provided by Cityscapes.
First, we compute the 3-dof transform between consecutive frames $t$ and $t+1$ from $o_t$ and $\Delta t_{t+1}$ 
by applying a velocity motion model~\cite{thrun2002probabilistic}, typically applied to a mobile robot.
More specifically,
\begin{align}
    \theta_t = \Delta\theta_t \cdot \Delta t_{t+1}, \\
    r_{t+1} = v_t / (\Delta\theta_t), \\
    x_{t+1} = r_{t+1} \cdot \sin{\theta_t}, \\ 
    y_{t+1} = r_{t+1} - r_{t+1} \cdot \cos{\theta_t},
\end{align}
where $(x_{t+1}, y_{t+1})$ is the location of the vehicle at time $t+1$ treating the vehicle at time $t$ as the origin (x-axis is front, y-axis is left, and z-axis is up), and $\theta_t$ is the rotation of the vehicle at time $t$ along the z-axis.
We then extend $(x, y, \theta)$ into a 6-dof transform $H^{t+1}_t$, and apply the camera extrinsics $H^{\text{cam}}_{\text{veh}}$ to obtain the transform of the cameras from frame $t$ to frame $t+1$ via,
\begin{align}
    H_{\text{veh}, t} = 
    \begin{bmatrix} 
        \cos{\theta_t} & -\sin{\theta_t} & 0 & x_{t+1} \\ 
        \sin{\theta_t} & \cos{\theta_t} & 0 & y_{t+1} \\ 
        0 & 0 & 1 & 0 \\ 
        0 & 0 & 0 & 1  \end{bmatrix}, \\
    H^{t+1}_t = H^{\text{cam}}_{\text{veh}} H_{\text{veh},t}^{-1} (H^{\text{cam}}_{\text{veh}})^{-1}.
\end{align}
The final transform $H_t^{T+F}$ from frame $t$ to frame $T+F$ is obtained by concatenating the consecutive transforms $H^{t+1}_t$ for $t \in \{1, \dots , T+F-1\}$. 

For egomotion estimation (\secref{sec:odom}), $o_t$ is used as input for frames $t \in \{1,\dots, T\}$ and predicted for frames $t\in\{T+1, \dots, T+F\}$. For `things' forecasting (\secref{sec:fg}), the input egomotion vector consists of $v_t$, $\theta_t$, $x_{t+1}$, $y_{t+1}$, and $\theta_{t+1}$, \ie, the speed and yaw rate of the vehicle as well as how far the vehicle moves in this time step. 

\section{Detail on background modelling}
\label{app:background}
Here we give more details on our background modelling, to ensure reproducibility.

\begin{figure}
    \includegraphics[width=\columnwidth]{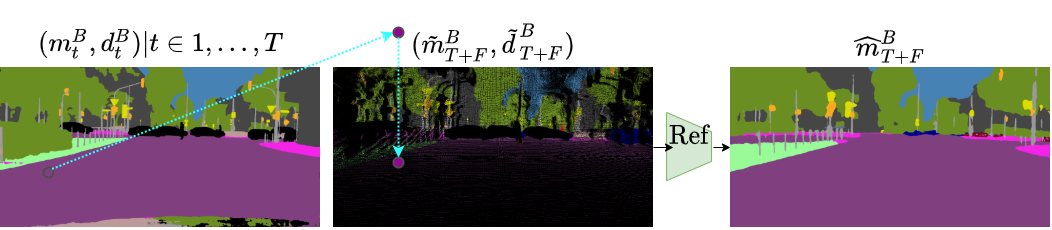}
    \vspace{-7pt}
    \caption{\textbf{`Stuff' forecasting is achieved in two steps.} First, for each input frame $t \in \{1, \dots, T\}$ given depth map $d^B_t$ and ego-motion, project the semantic map $m^B_t$ for all the background pixels on to frame $T+F$. Note the sparsity of the projected semantic maps. Second, we apply a refinement network to complete the background segmentation.}
    \label{fig:background}
\end{figure}

The background model is summarized in  \figref{fig:background}.
More formally, the background model estimates the semantic segmentation of an unseen future frame via
\begin{align}
    \widehat{m}^B_{T+F} = \text{ref}( \, \{ \, \text{proj}( m_t, d_t, K, H_t, u_t ) \, \}|_{t=1}^T \, ),
\end{align}
where $K$ subsumes the camera intrinsics, $H_t$ is the 6-dof camera transform from input frame $t$ to target frame $T+F$, $m_t$ is the semantic estimate at frame $t$ obtained from a pre-trained semantic segmentation model, $d_t$ is the input depth map at time $t$, and $u_t$ denotes the coordinates of all the pixels not considered by the foreground model.
Here, $\textit{proj}$ refers to the step that creates the sparse projected semantic map at frame $T+F$ from an input frame $t$, and $\textit{ref}$ refers to a refinement model that completes the background segmentation.

The refinement model $\textit{ref}$ receives a reprojected semantic point cloud $(\tilde{m}_t^B, \tilde{d}_t^B)$ from applying $\textit{proj}$ to each input frame at time $t\in\{1,\dots,T\}$. 
For $\textit{proj}$, each input frame comes with a per-pixel semantic label $m_t$ and a depth map $d_t$ obtained through a pre-trained model. We back-project, transform, and project the pixels from an input frame to the target frame. 
This process can be summarized as
\begin{align}
    \begin{bmatrix} x_t \\ y_t \\ z_t \end{bmatrix} = H_t \begin{bmatrix} K^{-1} \begin{bmatrix} u_t \\ 1 \end{bmatrix} D_t \\ 1 \end{bmatrix}, \\
    \begin{bmatrix} u_{T+F} \\ 1 \end{bmatrix} = K \begin{bmatrix}  x_t / z_t \\ y_t / z_t \\ 1 \end{bmatrix}, \\
    \tilde{m}_t^B(u_{T+F}) = m^B_t(u_t), \\
    \tilde{d}_t^B(u_{T+F}) = z_t,
\end{align}
where $u_t$ is a vector of the pixel locations for all background locations in the image at time $t$, $K$ subsumes the camera intrinsics, $D_t$ is a diagonal matrix whose entries contain depth $d_t$ of the corresponding pixels, and
$u_{T+F}$ is the vector containing corresponding pixel locations for the image at time $T+F$. 
We maintain the per-pixel semantic class  obtained from the frame $m_t$, and the projected depth. 
Note, in this process, if multiple pixels $u_t$ from an input frame are projected to the same pixel $u_{T+F}$ in the target frame, we keep the depth and semantic label of the one with the smallest depth value (\ie, closest to the camera).

\section{Foreground and background aggregation steps}
\label{app:aggregation}

\begin{algorithm}[t]
\caption{Foreground and background aggregation}
\label{alg:aggregation}
\begin{algorithmic}[1]
\STATE {\bfseries Input:} Background semantics $m^B_{T+F}$; \\
                  \quad\quad\quad Foreground segmentations $m^i_{T+F}$, classes $c^i$, \\
                  \quad\quad\quad depths $d^i_{T+F}, i=1, \dots, N$; \\
\FOR{$(x,y) \in \{1,\dots, W\} \times \{1, \dots, H\}$}
    \STATE $S_{T+F}(x, y) \leftarrow [m^B_{T+F}(x, y), 0]$
\ENDFOR
\STATE $\sigma \leftarrow \text{ArgSortDescending}(d^1_{T+F}, \dots, d^N_{T+F})$
\FOR{$i \in \sigma$}
    \FOR{$(x,y) \in \{1,\dots, W\} \times \{1, \dots, H\}$}
        \IF{$m^i_{T+F} = 1$}
            \STATE $S_{T+F}(x, y) \leftarrow [c^i, i]$
        \ENDIF
    \ENDFOR
\ENDFOR
\STATE {\bfseries Return:} future panoptic segmentation $S_{T+F}$
\end{algorithmic}
\end{algorithm}

\algref{alg:aggregation} describes our aggregation steps in detail. We initialize the output using the predicted background segmentation. After this, instances are sorted in reverse order of depth, and then they are placed one by one on top of the background.

\section{Losses}
\label{app:losses}
Here we formally describe  the losses used to train the model.

\subsection{`Things' forecasting loss}
\label{sec:sup_foreground_loss}
For   instance $i$ we use $\mathcal{L}_\text{fg}^i \coloneqq$ 
\begin{align}
     \hspace{-0.2cm}\frac{1}{Z_i} \!\sum_{t=T}^{T+F}\! p(t,i)\left(\lambda \text{SmoothL1}(\widehat{\mathbf{x}}_t^i, \mathbf{x}_t^i) + \text{MSE}(\widehat{\mathbf{r}}_t^i, \mathbf{r}_t^i)\right),
\end{align}
where $Z_i = \sum_{t=T}^{T+F}p(t,i)$ is a normalization constant and $p(t,i)$ equals $1$ if we have an estimate for instance $i$ in frame $t$ while being $0$ otherwise.  MSE refers to mean squared error, \ie, $\text{MSE}\left(\widehat{\mathbf{r}}_t^i, \mathbf{r}_t^i)\right) \coloneqq \frac{1}{J}\sum_{j=1}^J \left(\widehat{\mathbf{r}}_t^i- \mathbf{r}_t^i\right)^2$,  while SmoothL1 is given by 
\begin{align}
    \text{SmoothL1}(\textbf{a}, \textbf{b}) &\coloneqq \frac{1}{J}\sum_{j=1}^J \text{SmoothL1Fn}(\mathbf{a}^j, \mathbf{b}^j)\\
    \text{SmoothL1Fn}(a, b) &\coloneqq \begin{cases} \tfrac{1}{2}(a-b)^2, &\text{ if } |a-b| < 1,\\ 
    |a-b| - \tfrac{1}{2}, &\text{otherwise}, \end{cases}
\end{align}
where $\mathbf{a}$ and $\mathbf{b}$ are vector-valued inputs and $a$ and $b$ are scalars.
We use $\lambda=0.1$, which was chosen to balance the magnitudes of the two losses. 

\subsection{`Stuff' forecasting loss}
\label{sec:sup_background_loss}

To train the refinement network we use the cross-entropy loss 
\begin{align}
    \mathcal{L}_\text{bf} \coloneqq \tfrac{1}{\sum_{x,y}\mathbf{1}^{\text{bg}}_{t}[x,y]} \sum_{x,y} \mathbf{1}^\text{bg}_{t}[x,y] \sum_{c} m^{i*}_{t}(x,y,c) \log\left( p^i_t(x,y)\right). 
\end{align}
Here, $\mathbf{1}^{\text{bg}}_{t}[x,y]$ is an indicator function which specifies whether pixel coordinates $(x,y)$ are in the background of frame $t$, and $m_t^{i*}(x,y,c) = 1$ if $c$ is the correct class for pixel $(x,y)$ and $0$ otherwise.
Other variables are as described in the main paper.

\section{Further implementation details}
\label{app:impdetails}
\subsection{`Things' forecasting model} For instance detection, we use the pretrained MaskR-CNN model provided by Detectron2\footnote{\url{https://github.com/facebookresearch/detectron2}}, which is first trained on the COCO dataset~\cite{lin2014microsoft} and then finetuned on Cityscapes. For all detections, we extract the object bounding box and the $256\times14\times14$ feature tensor extracted after the ROIAlign stage. The detected instances are provided to DeepSort~\cite{wojke2017simple} to retrieve associations across time. We use a pre-trained model\footnote{\url{https://github.com/nwojke/deep_sort}} which was trained on the MOT16 dataset~\cite{milan2016mot16}. The tracker is run on every $30$ frame sequence once for every `things' class (in other words, the tracker is only asked to potentially link instances of the same class as determined by instance detection). 

Both $\text{GRU}_\text{enc}$ and $\text{GRU}_\text{dec}$ are $1$-layer GRUs with a hidden size of $128$. Both $\text{ConvLSTM}_\text{enc}$ and $\text{ConvLSTM}_\text{enc}$ are $2$-layer ConvLSTM networks using $3\times3$ kernels and $256$ hidden channels. $f_\text{enc,b}$ and $f_\text{bbox}$ are $2$-layer multilayer perceptrons with ReLU activations and a hidden size of $128$. $f_\text{bfeat}$ is a linear layer that produces a $16$-dimensional output, which is then copied across the spatial dimensions to form a $16 \times 14 \times 14$ tensor and concatenated with the mask feature tensor along the channel dimension before being used as input. $f_\text{mfeat}$ is a $1\times1$ convolutional layer producing a $8$ channel output, which is followed by a ReLU nonlinearity and a linear layer which produces a $64$-dimensional output vector. $f_\text{enc,m}$ and $f_\text{mask}$ are $1\times1$ convolutional layers producing a $256$ channel output.  MaskOut is the mask head from MaskRCNN, and consists of $4$ $3\times3$ convolutional layers with output channel number $256$, each followed by ReLU nonlinearities, a $2\times2$ ConvTranspose layer, and a final $1\times1$ convolutional layer that produces an $8$ channel output. Each channel represents the output for a different class, and the class provided as input is used to select the proper mask. The Mask Head's parameters are initialized from the same pre-trained model as used during instance detection and are fixed during training.

\begin{table*}[t]
    \vspace{-0.2cm}
    \footnotesize
      \centering
      \begin{tabular}{l ccc ccc ccc}
    \toprule

    & \multicolumn{3}{c}{All} & \multicolumn{3}{c}{Things} & \multicolumn{3}{c}{Stuff}\\
    & PQ & SQ & RQ & PQ & SQ & RQ & PQ & SQ & RQ \\
    \midrule
    Flow & $25.6$ & $70.1$ & $34.0$ & $12.4$ & $66.3$ & $18.1$ & $35.3$ & $72.9$ & $45.5$\\
    Hybrid \cite{terwilliger2019recurrent} (bg) and \cite{luc2018predicting} (fg) & $29.4$ & $69.8$ & $38.5$ & $18.0$ & $67.2$ & $25.7$& $37.6$ & $71.6$ & $47.8$\\
    \textbf{Ours} & $\textbf{35.7}$ & $\textbf{72.0}$ & $\textbf{46.5}$ & $\textbf{24.0}$ & $\textbf{69.0}$ & $\textbf{33.7}$ & $\textbf{44.2} $ & $\textbf{74.2}$ & $\textbf{55.8}$\\
    \bottomrule
      \end{tabular}
      \caption{\textbf{Panoptic segmentation forecasting evaluated on the Cityscapes test set, mid-term.}  
      Higher is better for all metrics.}
      \label{tab:panoptic_test}
 \end{table*}
 
During training, individual object tracks were sampled from every video sequence and from every $18$-frame sub-sequence from within each video. The tracking model we used frequently made ID switch errors for cars which the ego vehicle was driving past; specifically, the tracker would label a car visible to the camera as being the same car that the recording vehicle drove past a few frames previously. This had the effect of causing some tracks to randomly ``jump back'' into the frame after having left. In an attempt to mitigate these errors, instance tracks belonging to cars were truncated if this behavior was detected in an input sequence. Specifically, if a given car track, after being located within $250$ pixels from the left or right edge of a previous input frame, moved towards the center of the frame by more than $20$ pixels in the current frame, it was discarded for this and all future frames. Since this led to incomplete tracks in some cases, car tracks located within $200$ pixels from the left or right side of the frame were augmented by estimating their velocity from previous inputs and linearly extrapolating their locations until they were no longer present in the frame. 

We trained for $200000$ steps using a batch size of $32$, a learning rate of $0.0005$, and the Adam optimizer. Gradients were clipped to a norm of $5$. The learning rate was decayed by a factor of $0.1$ after $100000$ steps. During training, ground-truth odometry was used as input; odometry predictions were used for future frames for all evaluations except for the ablation which is listed as having used ground-truth future odometry. Bounding box features and odometry were normalized by their means and standard deviations as computed on the training data. During evaluation, we filtered all sequences which did not detect the object in the most recent input frame (\ie, $T=3$), as this led to improved performance.

\subsection{`Stuff' forecasting model} 
We use the model described by Zhu~\etal~\cite{zhu2019improving} with the SEResNeXt50 backbone and without label propagation as our single frame semantic segmentation model\footnote{\url{https://github.com/NVIDIA/semantic-segmentation/tree/sdcnet}}. For our refinement model, we use a fully convolutional variant of the HarDNet architecture~\cite{chao2019hardnet} which is set up to predict semantic segmentations\footnote{\url{https://github.com/PingoLH/FCHarDNet}}. We initialize with pre-trained weights and replace the first convolutional layer with one that accepts a $60$ channel input ($3$ input frames, each consisting of a $19$ channel 1-hot semantic input and a $1$ channel depth input). We trained a single model for both the short- and mid-term settings; hence, for each sequence in the training data, we use two samples: one consisting of the $5$th, $8$th, and $11$th frames (to represent the mid-term setting) and one consisting of the $11$th, $14$th, and $17$th frames (to represent the short-term setting), where frame $20$ was always the target frame.
During training, ground-truth odometry was used to create the point cloud transformations; during evaluation unless specified otherwise, the predicted odometry was used for future frames.

We trained the `stuff' forecasting model for $90000$ steps using a batch size of $16$, a learning rate of $0.002$, weight decay $0.0001$, and momentum $0.9$. Gradients were clipped to a norm of $5$. The learning rate was decayed by a factor of $0.1$ after $50000$ steps. During training, we randomly resized inputs and outputs between factors of $[0.5, 2]$ before taking $800\times800$ crops. Input depths were clipped to lie between $[0.1, 200]$ and normalized using mean and standard deviation computed on the training set.

 \begin{table}[t]

\footnotesize
  \centering
  \begin{tabular}{lcccc}
\toprule
& \multicolumn{2}{c}{Short term: $\Delta t = 3$} & \multicolumn{2}{c}{Mid term: $\Delta t = 9$} \\
Accuracy (mIoU) & All & MO & All & MO \\
\midrule
F2MF \cite{saric2020warp}$^*$ & $70.2$ & $68.7$ & $59.1$ & $56.3$\\
\textbf{Ours} & $67.3$ & $58.8$ & $57.7$ & $48.8$  \\
\bottomrule
  \end{tabular}
  \vspace{1pt}
  \caption{\textbf{Semantic segmentation forecasting results on the Cityscapes test dataset.}  Baseline numbers,  are from \cite{saric2020warp}; the * indicates training on both train and validation data. Higher is better for all metrics.}
  \label{tab:test_semantic}
 \end{table}

\subsection{Egomotion estimation} 
The egomotion estimation model takes as input $o_t \coloneqq [v_t, \theta_t], t \in \{1, \dots, T\}$, where $v_t$ is the speed and $\theta_t$ is the yaw rate of the ego-vehicle. It is tasked to predict $\widehat{o}_{T+1}, \dots, \widehat{o}_{T+F}$. Unlike the other components of our approach, the egomotion estimation model does not subsample inputs -- hence, it takes $9$ steps of input and predicts the odometry for the next $9$ steps. $\text{GRU}_\text{cam}$ is a $1$-layer GRU with a hidden size of $128$. $f_\text{cam}$ is a linear layer.

We trained the egomotion estimation model for $80000$ steps using a batch size of $32$, a learning rate of $0.0005$, and the Adam optimizer. Gradients were clipped to a norm of $5$. Inputs were normalized using means and standard deviations computed on the training set.

\subsection{Training and Inference time}
During inference, our unoptimized code makes a prediction in about 700ms using one 32GB NVIDIA V100. Additional engineering efforts can reduce this time further. Training the foreground and background models takes approximately 12 and 18 hours, respectively, on the same GPU.

\subsection{ORB-SLAM details} 
We ran ORB-SLAM3~\cite{ORBSLAM3_2020}\footnote{\url{https://github.com/UZ-SLAMLab/ORB_SLAM3}} with stereo images to obtain ego-motion in our ablation study. Each sequence (30 frames) is treated as its own SLAM session providing 6-dof poses for all the frames in the sequence. We then converted the poses into speed and yaw rate for each frame according to their timestamps.

\subsection{Monocular depth details}

Our experiments in Tab.~\ref{tab:ablations} show we can still achieve good performance even if we don't have stereo pairs at test time, through the use of a monocular depth estimation framework.
For this, we used the codebase of \cite{godard2019digging} to finetune a model for depth estimation on Cityscapes.

We initialised this model with the authors' weights trained on the KITTI dataset \cite{Geiger2012CVPR}.
We used their high-resolution Resnet-18 model, which was trained with a KITTI-image input size of $1024\times320$ pixels\footnote{Available from \url{https://github.com/nianticlabs/monodepth2}}.
We then finetuned this model on the training split of the Cityscapes dataset.
For Cityscapes finetuning we trained on the full Cityscapes input image without any cropping, and we trained using self-supervised reprojection losses using the stereo pairs as supervision (we did not use the monocular sequences, though we expect including these at training could further improve scores). 
We resized our Cityscapes input images to $640\times320$ pixels at both training and test time.
We used a smaller image input size than the model was trained for on KITTI to account for the change in image aspect ratio between KITTI and Cityscapes.
We trained our model with Adam \cite{adamsolver} for 15 epochs with a learning rate of $1e-4$ and then 15 epochs with $1e-5$.

\begin{table}[t]
\footnotesize
\centering
\begin{tabular}{lcccc}
\toprule
&\multicolumn{2}{c}{Short term: $\Delta t = 3$} & \multicolumn{2}{c}{Mid term: $\Delta t = 9$} \\
& AP & AP50 & AP & AP50 \\
\midrule
F2F \cite{luc2018predicting} & - & - & $6.7$ & $17.5$ \\
Ours & $14.9$ & $31.3$ & $\textbf{8.4}$ & $\textbf{19.8}$\\
\bottomrule
\end{tabular}
\caption{\textbf{Instance segmentation forecasting on the Cityscapes Test dataset.} Higher is better for all metrics.}
\label{tab:inst_test}
\end{table}
\subsection{Flow Baseline}
To implement this model, we warp the panoptic segmentation computed by the oracle model on the most recently seen input frame using optical flow. Specifically, we compute the forward optical flow between the most recent input frames, \ie, between frames for $t\in\{2,3\}$, using the CSS configuration of the model introduced by Ilg~\etal~\cite{ilg2017flownet}\footnote{\url{https://github.com/NVIDIA/flownet2-pytorch}}. We then iteratively warp the inputs via the following procedure: first, we warp the optical flow using itself (to align it with the current frame), and then we warp the current panoptic segmentation using the resulting flow. For short-term, this is done once, and for mid-term, this is done three times. This was the best performing procedure we tried. Negating reverse optical flow, extrapolating the flow linearly for the mid-term setting, or a combination of the two resulted in worse performance.

\subsection{Hybrid Semantic/Instance Forecasting Model}
The semantic segmentation forecasting component of the hybrid baseline consists of the model developed by Terwilliger~\etal~\cite{terwilliger2019recurrent}, which anticipates the optical flow between the most recently seen input frame and the target frame and uses it to warp the predicted semantic segmentation for that input frame. The authors originally evaluated on four settings: $\Delta t = 1$, $\Delta t = 3$, $\Delta t = 9$, and $\Delta t = 10$. Their official implementation\footnote{\url{https://github.com/adamtwig/segpred}} provides pre-trained models for $\Delta t = 3$ and $\Delta t = 10$; these are what we use for our short- and mid-term settings, respectively. Note,
that Terwilliger~\etal~\cite{terwilliger2019recurrent} don't provide a model for $\Delta t = 9$ or the code they used to train their models. 
Despite the fact that the $\Delta t=10$ setting is marginally more challenging than the $\Delta t=9$ setting, Terwilliger~\etal report better performance on the $\Delta t=10$ setting \cite{terwilliger2019recurrent}, which is why we felt comfortable using this approach for our hybrid model. The best performing semantic segmentation forecasting approach, developed by \v Sari\'c~\etal~\cite{saric2020warp}, did not have code or models available at time of submission.

The instance segmentation forecasting component of the hybrid baseline consists of the model developed by Luc~\etal~\cite{luc2018predicting}, which anticipates the image features for the target frame autoregressively. We use their official implementation\footnote{\url{https://github.com/facebookresearch/instpred}}, which provides pre-trained models for both the short- and mid-term settings. 

Predictions are fused by `pasting' all `thing' class predictions made by the instance segmentation forecasting model on top of the `stuff' class predictions made by the semantic segmentation forecasting model. Note that this leaves some pixels without predictions; this occurs for pixels where the semantic forecasting model predicted a `thing' class but the instance forecasting model did not predict a `thing' class. We implemented a version of this model which filled in these gaps with the closest `stuff' class predicted by the semantic forecasting model, but this performed worse.

\begin{table}
\setlength{\tabcolsep}{3pt}
    \footnotesize
      \centering
      \begin{tabular}{lcccccc}
    \toprule
    & \multicolumn{3}{c}{$\Delta t = 3$} & \multicolumn{3}{c}{$\Delta t = 9$} \\
    & PQ & SQ & RQ & PQ & SQ & RQ \\
    \midrule
    \textbf{Ours} & $\textbf{49.0}$ & $74.9$ & $\textbf{63.3}$ & $\textbf{36.3}$ & $71.3$ & $\textbf{47.8}$ \\
    a) fg w/ independent RNNs & $\textbf{49.2}$ & $\textbf{75.2}$ & $63.3$ & $35.7$ & $71.2$ & $46.9$ \\
    b) fg w/o velocity features & $\textbf{49.2}$  & $75.0$  & $\textbf{63.4}$ & $35.7$ & $71.2$ & $47.0$ \\
    \bottomrule
      \end{tabular}
      \caption{\textbf{Additional ablations} on Panoptic segmentation forecasting using Cityscapes. Higher is better for all metrics. All approaches use predicted future odometry.}
      \label{tab:appx_ablations}
 \end{table}

\section{More Experimental Results}
\label{sec:appx_eval}
\begin{table*}[t]
    \centering
    \newcommand*\rot{\rotatebox{90}}
    
    \resizebox{1.0\textwidth}{!}{
    \begin{tabular}{l|ccccccccccccccccccc|c}
    \toprule
        & \rot{road} & \rot{sidewalk} & \rot{building} & \rot{wall} & \rot{fence} & \rot{pole} & \rot{traffic light} & \rot{traffic sign} & \rot{vegetation} & \rot{terrain} & \rot{sky} & \rot{person} & \rot{rider} & \rot{car} & \rot{truck} & \rot{bus} & \rot{train} & \rot{motorcycle} & \rot{bicycle} & \rot{mean} \\
        \midrule
        \textbf{Deeplab (Oracle)$\dagger$} & $97.9$ & $78.2$ & $88.5$ & $29.4$ & $38.9$ & $60.0$ & $55.6$ & $74.5$ & $89.5$ & $36.1$ & $87.9$ & $50.8$ & $46.4$ & $67.3$ & $51.5$ & $66.6$ & $37.8$ & $44.2$ & $44.1$ & $60.3$\\
        \midrule
        \textbf{Deeplab (Last seen frame)} & $94.3$ & $52.4$ & $71.1$ & $11.3$ & $19.4$ & $6.1$ & $12.9$ & $15.0$ & $72.1$ & $16.9$ & $72.7$ & $10.3$ & $8.0$ & $29.6$ & $35.1$ & $51.7$ & $24.2$ & $9.8$ & $7.9$ & $32.7$ \\
        Flow & $95.6$ & $61.5$ & $79.8$ & $17.3$ & $\textbf{28.6}$ & $8.7$ & $26.2$ & $36.8$ & $80.7$ & $\textbf{26.9}$ & $79.7$ & $21.0$ & $14.0$ & $43.4$ & $40.6$ & $56.8$ & $26.7$ & $23.2$ & $18.7$ & $41.4$\\
        Hybrid \cite{terwilliger2019recurrent} (bg) and \cite{luc2018predicting} (fg) & $\textbf{96.2}$ & $63.4$ & $81.4$ & $23.1$ & $23.7$ & $7.1$ & $19.1$ & $36.9$ & $82.3$ & $20.3$ & $79.8$ & $26.8$ & $21.8$ & $46.4$ & $\textbf{42.2}$ & $60.0$ & $41.4$ & $25.6$ & $22.5$ & $43.2$\\
        \textbf{Ours} & $\textbf{96.2}$ & $\textbf{66.1}$ & $\textbf{83.5}$ & $\textbf{26.1}$ & $27.4$ & $\textbf{31.7}$ & $\textbf{37.0}$ & $\textbf{49.9}$ & $\textbf{84.8}$ & $26.1$ & $\textbf{82.0}$ & $\textbf{31.8}$ & $\textbf{31.5}$ & $\textbf{48.8}$ & $\textbf{42.2}$ & $\textbf{61.2}$ & $\textbf{47.0}$ & $\textbf{31.4}$ & $\textbf{27.3}$ & $\textbf{49.0}$\\
        \bottomrule
    \end{tabular}}
    
    \vspace{5pt}
    \caption{\textbf{Per-class results for Panoptic Quality on Cityscapes validation dataset (short-term).} }
    \label{tab:per_class_pq_short}
\end{table*}
\begin{table*}[t]
    \centering
    \newcommand*\rot{\rotatebox{90}}
    
    \resizebox{1.0\textwidth}{!}{
    \begin{tabular}{l|ccccccccccccccccccc|c}
    \toprule
        & \rot{road} & \rot{sidewalk} & \rot{building} & \rot{wall} & \rot{fence} & \rot{pole} & \rot{traffic light} & \rot{traffic sign} & \rot{vegetation} & \rot{terrain} & \rot{sky} & \rot{person} & \rot{rider} & \rot{car} & \rot{truck} & \rot{bus} & \rot{train} & \rot{motorcycle} & \rot{bicycle} & \rot{mean} \\
        \midrule
        \textbf{Deeplab (Oracle)$\dagger$} & $97.9$ & $78.2$ & $88.5$ & $29.4$ & $38.9$ & $60.0$ & $55.6$ & $74.5$ & $89.5$ & $36.1$ & $87.9$ & $50.8$ & $46.4$ & $67.3$ & $51.5$ & $66.6$ & $37.8$ & $44.2$ & $44.1$ & $60.3$ \\
        \midrule
        \textbf{Deeplab (Last seen frame)} & $90.4$ & $32.5$ & $57.6$ & $7.6$ & $10.6$ & $4.6$ & $8.9$ & $7.4$ & $55.1$ & $8.8$ & $57.3$ & $5.3$ & $2.5$ & $13.2$ & $19.2$ & $27.3$ & $10.1$ & $4.7$ & $3.0$ & $22.4$  \\
        Flow & $90.5$ & $35.8$ & $66.2$ & $7.7$ & $15.0$ & $4.6$ & $11.9$ & $11.1$ & $65.6$ & $11.6$ & $64.4$ & $5.9$ & $2.5$ & $19.0$ & $21.5$ & $27.7$ & $13.5$ & $11.8$ & $5.3$ & $25.9$\\
        Hybrid \cite{terwilliger2019recurrent} (bg) and \cite{luc2018predicting} (fg) & $93.2$ & $44.9$ & $70.5$ & $12.4$ & $14.8$ & $1.2$ & $8.0$ & $10.8$ & $69.7$ & $13.9$ & $67.2$ & $8.0$ & $4.5$ & $27.3$ & $33.5$ & $41.7$ & $27.9$ & $8.3$ & $6.1$ & $29.7$\\
        \textbf{Ours} & $\textbf{93.9}$ & $\textbf{50.8}$ & $\textbf{76.4}$ & $\textbf{18.2}$ & $\textbf{19.9}$ & $\textbf{8.7}$ & $\textbf{18.7}$ & $\textbf{28.5}$ & $\textbf{77.0}$ & $\textbf{18.6}$ & $\textbf{72.7}$ & $\textbf{16.2}$ & $\textbf{12.0}$ & $\textbf{33.3}$ & $\textbf{36.1}$ & $\textbf{53.0}$ & $\textbf{29.8}$ & $\textbf{14.1}$ & $\textbf{12.6}$ & $\textbf{36.3}$\\
        \bottomrule
    \end{tabular}}
    
    \vspace{5pt}
    \caption{\textbf{Per-class results for Panoptic Quality on Cityscapes validation dataset (mid-term).} }
    \label{tab:per_class_pq_mid}
\end{table*}
\begin{table*}[t]
    \centering
    \newcommand*\rot{\rotatebox{90}}
    
    \resizebox{1.0\textwidth}{!}{
    \begin{tabular}{l|ccccccccccccccccccc|c}
    \toprule
        & \rot{road} & \rot{sidewalk} & \rot{building} & \rot{wall} & \rot{fence} & \rot{pole} & \rot{traffic light} & \rot{traffic sign} & \rot{vegetation} & \rot{terrain} & \rot{sky} & \rot{person} & \rot{rider} & \rot{car} & \rot{truck} & \rot{bus} & \rot{train} & \rot{motorcycle} & \rot{bicycle} & \rot{mean} \\
        \midrule
        \textbf{Deeplab (Oracle)$\dagger$} & $98.0$ & $85.6$ & $90.5$ & $74.3$ & $74.8$ & $69.7$ & $73.5$ & $80.1$ & $90.9$ & $75.7$ & $92.6$ & $76.0$ & $70.8$ & $84.2$ & $88.4$ & $90.8$ & $87.6$ & $73.8$ & $72.1$ & $81.5$ \\
        \midrule
        \textbf{Deeplab (Last seen frame)} & $94.4$ & $71.5$ & $78.8$ & $65.4$ & $65.6$ & $\textbf{67.0}$ & $\textbf{68.3}$ & $67.4$ & $77.8$ & $67.7$ & $83.0$ & $64.4$ & $60.1$ & $69.2$ & $74.7$ & $76.7$ & $75.7$ & $62.7$ & $63.4$ & $71.3$ \\
        Flow & $95.6$ & $76.0$ & $83.2$ & $68.5$ & $68.3$ & $65.0$ & $65.9$ & $67.3$ & $83.4$ & $69.1$ & $86.6$ & $65.6$ & $61.4$ & $75.8$ & $77.5$ & $80.0$ & $74.1$ & $66.1$ & $64.4$ & $73.4$\\
        Hybrid \cite{terwilliger2019recurrent} (bg) and \cite{luc2018predicting} (fg) & $\textbf{96.3}$ & $\textbf{77.2}$ & $84.9$ & $70.0$ & $69.0$ & $59.5$ & $63.6$ & $65.9$ & $84.6$ & $70.8$ & $86.5$ & $66.8$ & $61.9$ & $77.2$ & $\textbf{80.3}$ & $\textbf{83.1}$ & $\textbf{80.5}$ & $65.6$ & $63.8$ & $74.1$\\
        \textbf{Ours} & $\textbf{96.3}$ & $77.0$ & $\textbf{86.3}$ & $\textbf{71.1}$ & $\textbf{69.4}$ & $61.4$ & $65.4$ & $\textbf{70.6}$ & $\textbf{86.6}$ & $\textbf{71.3}$ & $\textbf{88.3}$ & $\textbf{67.7}$ & $\textbf{63.8}$ & $\textbf{77.7}$ & $81.4$ & $81.4$ & $74.8$ & $\textbf{67.6}$ & $\textbf{65.8}$ & $\textbf{74.9}$\\
        \bottomrule
    \end{tabular}}
    
    \vspace{5pt}
    \caption{\textbf{Per-class results for Segmentation Quality on Cityscapes validation dataset (short-term).} }
    \label{tab:per_class_sq_short}
\end{table*}
\begin{table*}[t]
    \centering
    \newcommand*\rot{\rotatebox{90}}
    
    \resizebox{1.0\textwidth}{!}{
    \begin{tabular}{l|ccccccccccccccccccc|c}
    \toprule
        & \rot{road} & \rot{sidewalk} & \rot{building} & \rot{wall} & \rot{fence} & \rot{pole} & \rot{traffic light} & \rot{traffic sign} & \rot{vegetation} & \rot{terrain} & \rot{sky} & \rot{person} & \rot{rider} & \rot{car} & \rot{truck} & \rot{bus} & \rot{train} & \rot{motorcycle} & \rot{bicycle} & \rot{mean} \\
        \midrule
        \textbf{Deeplab (Oracle)$\dagger$} & $98.0$ & $85.6$ & $90.5$ & $74.3$ & $74.8$ & $69.7$ & $73.5$ & $80.1$ & $90.9$ & $75.7$ & $92.6$ & $76.0$ & $70.8$ & $84.2$ & $88.4$ & $90.8$ & $87.6$ & $73.8$ & $72.1$ & $81.5$ \\
        \midrule
        \textbf{Deeplab (Last seen frame)} & $90.7$ & $68.2$ & $72.6$ & $63.4$ & $62.4$ & $\textbf{66.1}$ & $\textbf{72.7}$ & $\textbf{73.0}$ & $71.2$ & $64.0$ & $77.3$ & $63.7$ & $61.3$ & $66.8$ & $62.9$ & $70.8$ & $74.3$ & $56.4$ & $\textbf{64.4}$ & $68.5$ \\
        Flow & $90.8$ & $68.6$ & $76.0$ & $66.1$ & $64.1$ & $64.1$ & $69.0$ & $67.2$ & $75.0$ & $64.5$ & $78.5$ & $63.5$ & $60.4$ & $69.1$ & $70.2$ & $74.3$ & $\textbf{75.8}$ & $60.2$ & $63.0$ & $69.5$\\
        Hybrid \cite{terwilliger2019recurrent} (bg) and \cite{luc2018predicting} (fg) & $93.3$ & $69.7$ & $77.9$ & $66.6$ & $65.3$ & $59.9$ & $62.9$ & $61.9$ & $76.9$ & $65.1$ & $79.6$ & $63.7$ & $58.4$ & $71.5$ & $72.6$ & $72.2$ & $73.7$ & $62.1$ & $60.6$ & $69.1$\\
        \textbf{Ours} & $\textbf{94.1}$ & $\textbf{71.3}$ & $\textbf{81.5}$ & $\textbf{68.4}$ & $\textbf{66.8}$ & $59.0$ & $64.1$ & $65.1$ & $\textbf{80.9}$ & $\textbf{68.1}$ & $\textbf{83.0}$ & $\textbf{64.3}$ & $\textbf{61.4}$ & $\textbf{73.4}$ & $\textbf{76.9}$ & $\textbf{76.1}$ & $74.4$ & $\textbf{62.5}$ & $62.9$ & $\textbf{71.3}$ \\
        \bottomrule
    \end{tabular}}
    
    \vspace{5pt}
    \caption{\textbf{Per-class results for Segmentation Quality on Cityscapes validation dataset (mid-term).} }
    \label{tab:per_class_sq_mid}
\end{table*}
\begin{table*}[t]
    \centering
    \newcommand*\rot{\rotatebox{90}}
    
    \resizebox{1.0\textwidth}{!}{
    \begin{tabular}{l|ccccccccccccccccccc|c}
    \toprule
        & \rot{road} & \rot{sidewalk} & \rot{building} & \rot{wall} & \rot{fence} & \rot{pole} & \rot{traffic light} & \rot{traffic sign} & \rot{vegetation} & \rot{terrain} & \rot{sky} & \rot{person} & \rot{rider} & \rot{car} & \rot{truck} & \rot{bus} & \rot{train} & \rot{motorcycle} & \rot{bicycle} & \rot{mean} \\
        \midrule
        \textbf{Deeplab (Oracle)$\dagger$} & $99.9$ & $91.3$ & $97.8$ & $39.5$ & $52.1$ & $86.1$ & $75.6$ & $93.0$ & $98.5$ & $47.7$ & $94.9$ & $66.9$ & $65.5$ & $80.0$ & $58.2$ & $73.4$ & $43.1$ & $59.9$ & $61.2$ & $72.9$ \\
        \midrule
        \textbf{Deeplab (Last seen frame)} & $\textbf{99.9}$ & $73.4$ & $90.2$ & $17.3$ & $29.7$ & $9.1$ & $18.9$ & $22.2$ & $92.7$ & $25.0$ & $87.6$ & $16.0$ & $13.2$ & $42.8$ & $47.0$ & $67.4$ & $32.0$ & $15.6$ & $12.4$ & $42.7$  \\
        Flow & $\textbf{99.9}$ & $81.0$ & $95.9$ & $25.2$ & $\textbf{41.9}$ & $13.4$ & $39.7$ & $54.7$ & $96.8$ & $\textbf{39.0}$ & $92.0$ & $32.1$ & $22.8$ & $57.3$ & $52.4$ & $71.0$ & $36.0$ & $35.1$ & $29.0$ & $53.4$\\
        Hybrid \cite{terwilliger2019recurrent} (bg) and \cite{luc2018predicting} (fg) & $\textbf{99.9}$ & $82.1$ & $95.8$ & $33.0$ & $34.4$ & $11.9$ & $30.0$ & $56.0$ & $97.3$ & $28.8$ & $92.2$ & $40.1$ & $35.3$ & $60.2$ & $\textbf{52.6}$ & $72.2$ & $51.4$ & $39.1$ & $35.3$ & $55.1$\\
        \textbf{Ours} & $\textbf{99.9}$ & $\textbf{85.8}$ & $\textbf{96.7}$ & $\textbf{36.7}$ & $39.4$ & $\textbf{51.6}$ & $\textbf{56.6}$ & $\textbf{70.7}$ & $\textbf{97.9}$ & $36.6$ & $\textbf{92.9}$ & $\textbf{47.0}$ & $\textbf{49.3}$ & $\textbf{62.9}$ & $51.9$ & $\textbf{75.2}$ & $\textbf{62.9}$ & $\textbf{46.3}$ & $\textbf{41.5}$ & $\textbf{63.3}$\\
        \bottomrule
    \end{tabular}}
    
    \vspace{5pt}
    \caption{\textbf{Per-class results for Recognition Quality on Cityscapes validation dataset (short-term).} }
    \label{tab:per_class_rq}
\end{table*}
\begin{table*}[t]
    \centering
    \newcommand*\rot{\rotatebox{90}}
    
    \resizebox{1.0\textwidth}{!}{
    \begin{tabular}{l|ccccccccccccccccccc|c}
    \toprule
        & \rot{road} & \rot{sidewalk} & \rot{building} & \rot{wall} & \rot{fence} & \rot{pole} & \rot{traffic light} & \rot{traffic sign} & \rot{vegetation} & \rot{terrain} & \rot{sky} & \rot{person} & \rot{rider} & \rot{car} & \rot{truck} & \rot{bus} & \rot{train} & \rot{motorcycle} & \rot{bicycle} & \rot{mean} \\
        \midrule
        \textbf{Deeplab (Oracle)$\dagger$} & $99.9$ & $91.3$ & $97.8$ & $39.5$ & $52.1$ & $86.1$ & $75.6$ & $93.0$ & $98.5$ & $47.7$ & $94.9$ & $66.9$ & $65.5$ & $80.0$ & $58.2$ & $73.4$ & $43.1$ & $59.9$ & $61.2$ & $72.9$ \\
        \midrule
        \textbf{Deeplab (Last seen frame)} & $99.7$ & $47.6$ & $79.3$ & $12.1$ & $17.0$ & $7.0$ & $12.2$ & $10.1$ & $77.4$ & $13.7$ & $74.1$ & $8.3$ & $4.2$ & $19.8$ & $30.6$ & $38.5$ & $13.6$ & $8.3$ & $4.7$ & $30.4$ \\
        Flow & $99.7$ & $52.2$ & $87.1$ & $11.7$ & $23.4$ & $7.2$ & $17.3$ & $16.5$ & $87.4$ & $18.0$ & $82.1$ & $9.2$ & $4.2$ & $27.5$ & $30.6$ & $37.3$ & $17.8$ & $19.6$ & $8.4$ & $34.6$\\
        Hybrid \cite{terwilliger2019recurrent} (bg) and \cite{luc2018predicting} (fg) & $\textbf{99.9}$ & $64.5$ & $90.4$ & $18.7$ & $22.7$ & $2.1$ & $12.7$ & $17.4$ & $90.5$ & $21.4$ & $84.4$ & $12.5$ & $7.7$ & $38.2$ & $46.2$ & $57.9$ & $37.8$ & $13.4$ & $10.1$ & $39.4$\\
        \textbf{Ours} & $99.7$ & $\textbf{71.2}$ & $\textbf{93.7}$ & $\textbf{26.6}$ & $\textbf{29.8}$ & $\textbf{14.7}$ & $\textbf{29.2}$ & $\textbf{43.8}$ & $\textbf{95.1}$ & $\textbf{27.3}$ & $\textbf{87.6}$ & $\textbf{25.2}$ & $\textbf{19.5}$ & $\textbf{45.4}$ & $\textbf{47.0}$ & $\textbf{69.6}$ & $\textbf{40.0}$ & $\textbf{22.6}$ & $\textbf{20.0}$ & $\textbf{47.8}$\\
        \bottomrule
    \end{tabular}}
    
    \vspace{5pt}
    \caption{\textbf{Per-class results for Recognition Quality on Cityscapes validation dataset (mid-term)}. }
    \label{tab:per_class_rq_mid}
\end{table*}

\tabref{tab:panoptic_test} provides the panoptic segmentation forecasting metrics computed on the Cityscapes test dataset for our method as well as the flow and hybrid baselines for the mid-term setting. We outperform both other approaches on this data as well for all metrics. 

\tabref{tab:test_semantic} shows the semantic segmentation forecasting metrics computed on the Cityscapes test dataset for our approach as well as F2MF~\cite{saric2020warp}. F2MF outperforms our approach on the test data; however, note that the F2MF model used for test evaluation was trained on both the training and validation dataset, while we only train our model on the training dataset. Moreover, F2MF does not predict instance IDs for `thing' classes, meaning that this model cannot be used for panoptic segmentation.

\tabref{tab:inst_test} provides the instance segmentation forecasting metrics computed on the Cityscapes test dataset for our approach as well as F2F~\cite{luc2018predicting}. Our approach outperforms F2F on the mid-term setting. Luc~\etal do not provide an evaluation on the short-term setting, so we cannot directly compare against those results here.

\tabref{tab:appx_ablations} presents a few additional ablations analyzing the impact of our modeling choices: a)~\emph{fg w/ independent RNNs} `disconnects' the bounding box and appearance mask RNNs in the encoder and decoder so they do not use the outputs of $f_\text{mfeat}$ and $f_\text{bfeat}$ as input, and  b)~\emph{fg w/o velocity features} only uses location features as input, \ie, $\mathbf{x}^i_t \coloneqq [cx, cy, w, h, d]$. a) shows that joint modeling of instance motion and appearance mask leads to better performance at longer timescales. b) shows that guiding the network with temporal information is also important to improve longer-term forecasting.

Tables \ref{tab:per_class_pq_short}-\ref{tab:per_class_rq_mid} show the per-class breakdown of all panoptic segmentation metrics presented in \tabref{tab:panoptic}. The results shown in \tabref{tab:panoptic} consist of the average metric values computed over the set of classes in Cityscapes. For most classes, we outperform all other approaches on panoptic quality and recognition quality for both short- and mid-term. We additionally outperform all other approaches for many classes on segmentation quality, and outperform all other approaches on average.

\section{Visualization}
\label{sec:appx_viz}
\begin{figure*}[t]
    \centering
    \footnotesize
    \setlength\tabcolsep{0.5pt}
    \renewcommand{\arraystretch}{0.5}
    \begin{tabular}{ccccc}
        Last Seen Image &  Oracle & Flow & Hybrid & Ours\\
         \includegraphics[width=0.19\textwidth]{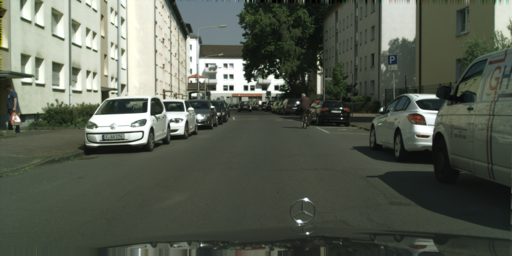}&
         \includegraphics[width=0.19\textwidth]{figs/viz/frankfurt_000000_000576_oracle.png}& \includegraphics[width=0.19\textwidth]{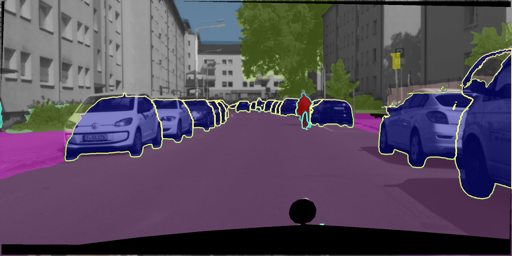}& \includegraphics[width=0.19\textwidth]{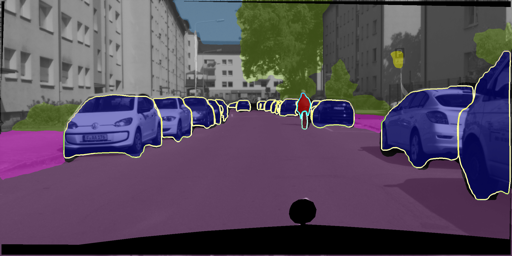}& \includegraphics[width=0.19\textwidth]{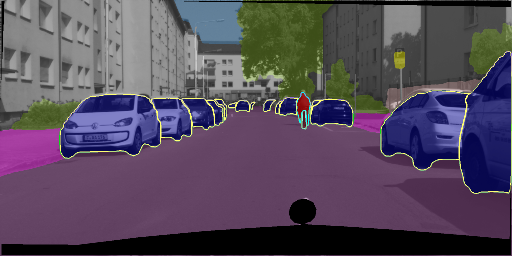}\\
         \includegraphics[width=0.19\textwidth]{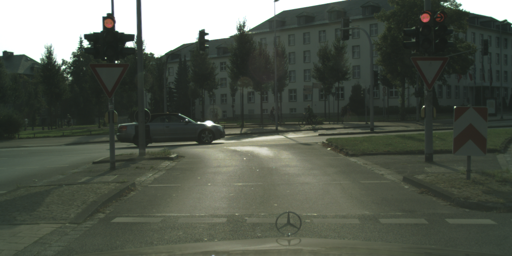}&
         \includegraphics[width=0.19\textwidth]{figs/viz/munster_000159_000019_oracle.png}& \includegraphics[width=0.19\textwidth]{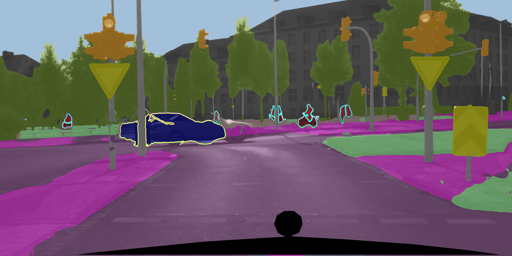}& \includegraphics[width=0.19\textwidth]{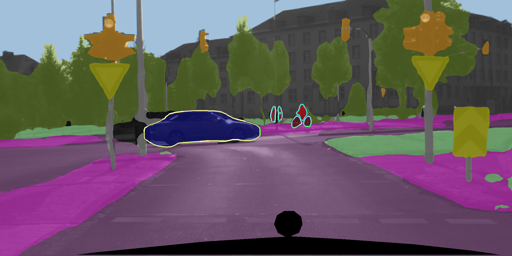}& \includegraphics[width=0.19\textwidth]{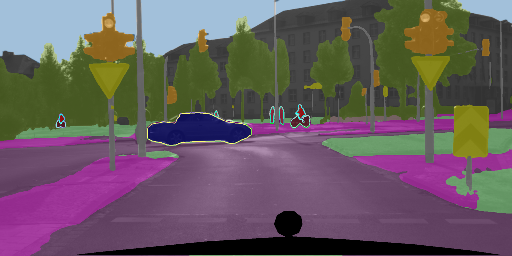}\\
         \includegraphics[width=0.19\textwidth]{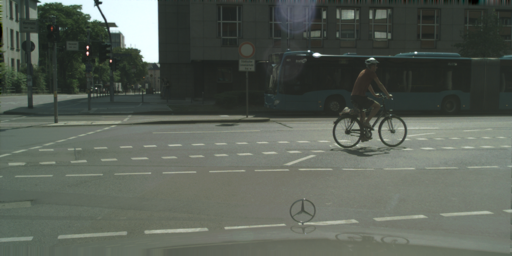}&
         \includegraphics[width=0.19\textwidth]{figs/viz/frankfurt_000001_001464_oracle.png}& \includegraphics[width=0.19\textwidth]{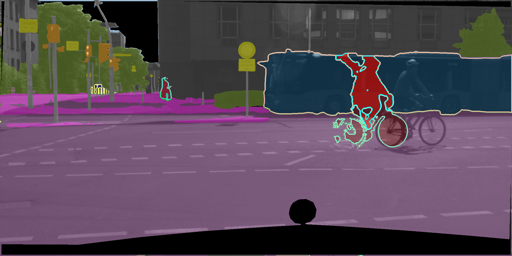}& \includegraphics[width=0.19\textwidth]{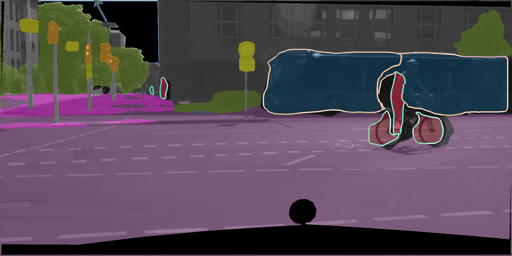}& \includegraphics[width=0.19\textwidth]{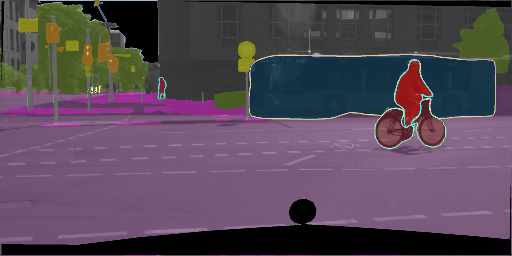}\\
         \includegraphics[width=0.19\textwidth]{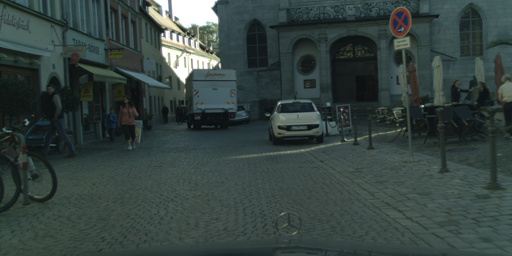}&
         \includegraphics[width=0.19\textwidth]{figs/viz/lindau_000024_000019_oracle.png}& \includegraphics[width=0.19\textwidth]{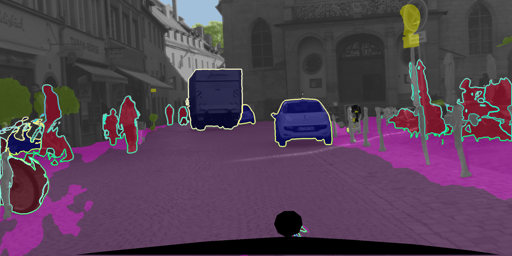}& \includegraphics[width=0.19\textwidth]{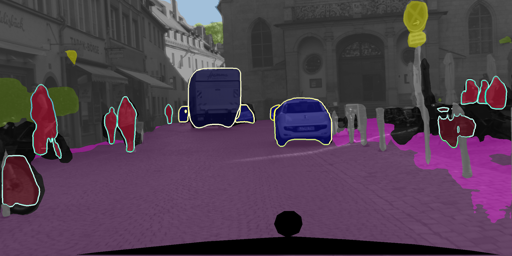}& \includegraphics[width=0.19\textwidth]{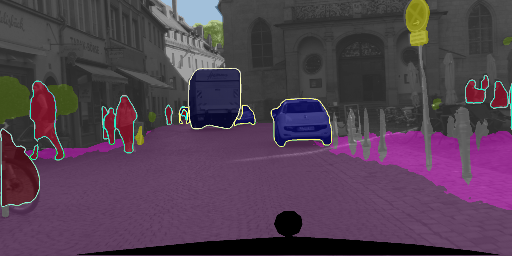}\\
    \end{tabular}
    \caption{\textbf{Short-term panoptic segmentation forecasts on Cityscapes.}}
    \label{fig:panoptic_viz_short}
    \vspace{-0.4cm}
\end{figure*}

\begin{figure*}[t]
    \centering
    \footnotesize
    \setlength\tabcolsep{0.5pt}
    \renewcommand{\arraystretch}{0.5}
    \begin{tabular}{ccccc}
        Last Seen Image &  Oracle & Flow & Hybrid & Ours\\
         \includegraphics[width=0.19\textwidth]{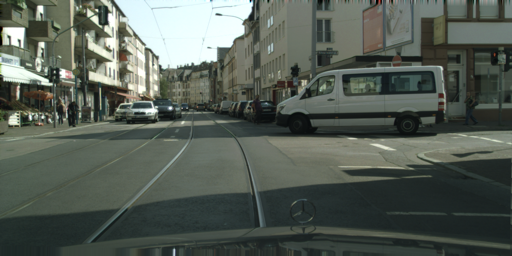}&
         \includegraphics[width=0.19\textwidth]{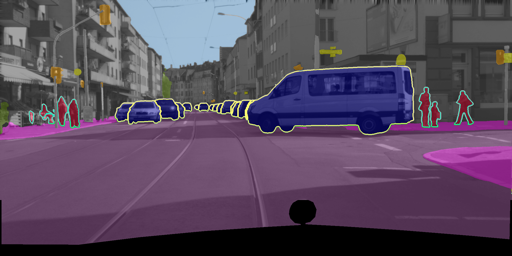}& \includegraphics[width=0.19\textwidth]{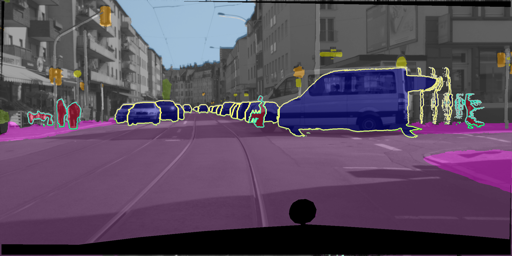}& \includegraphics[width=0.19\textwidth]{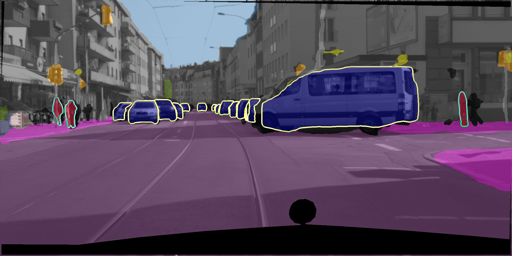}& \includegraphics[width=0.19\textwidth]{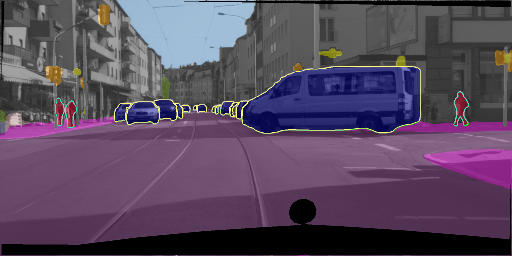}\\
         \includegraphics[width=0.19\textwidth]{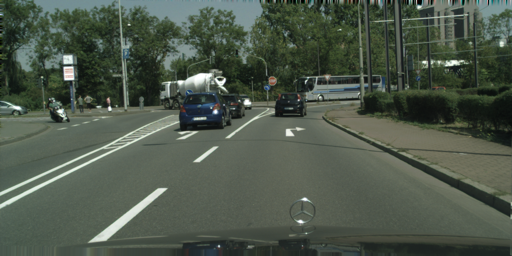}&
         \includegraphics[width=0.19\textwidth]{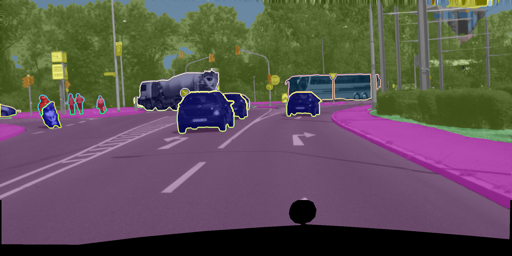}& \includegraphics[width=0.19\textwidth]{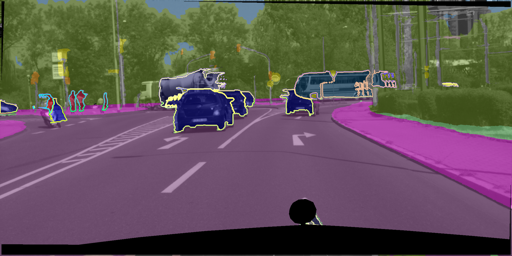}& \includegraphics[width=0.19\textwidth]{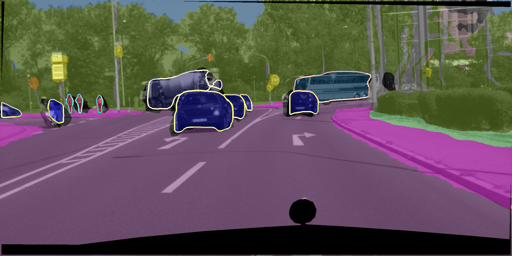}& \includegraphics[width=0.19\textwidth]{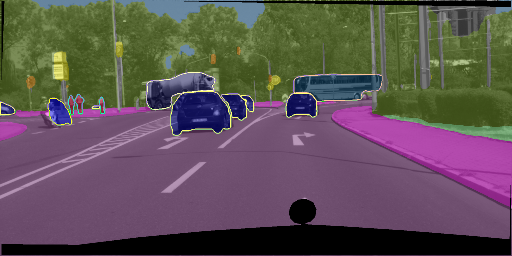}\\
         \includegraphics[width=0.19\textwidth]{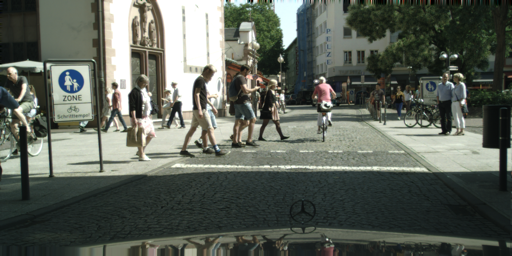}&
         \includegraphics[width=0.19\textwidth]{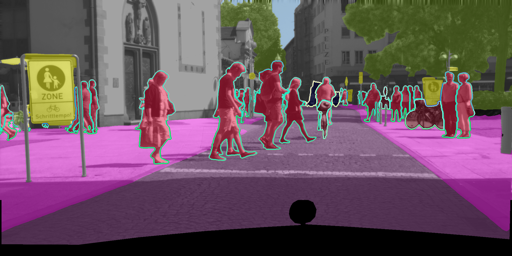}& \includegraphics[width=0.19\textwidth]{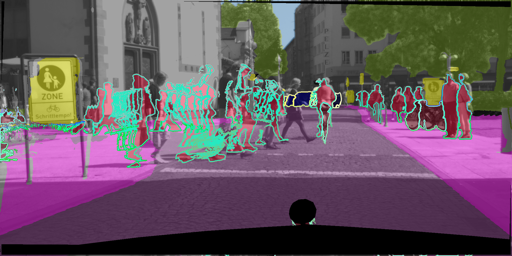}& \includegraphics[width=0.19\textwidth]{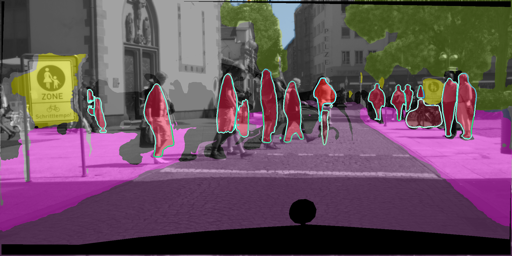}& \includegraphics[width=0.19\textwidth]{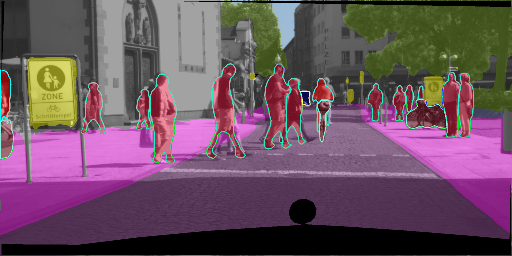}\\
         \includegraphics[width=0.19\textwidth]{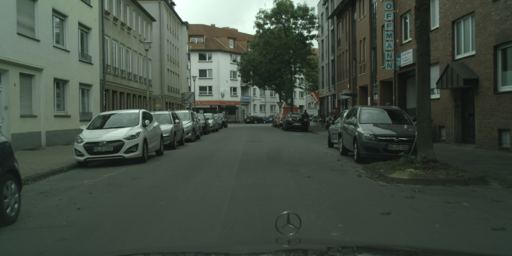}&
         \includegraphics[width=0.19\textwidth]{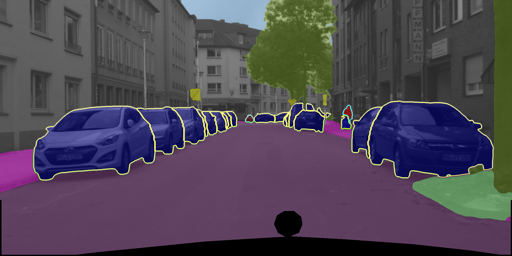}& \includegraphics[width=0.19\textwidth]{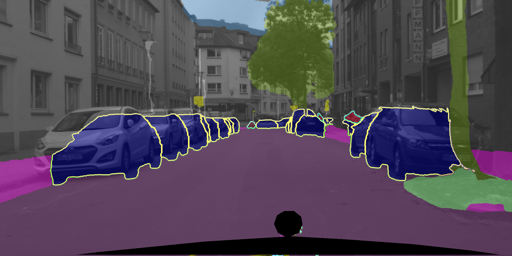}& \includegraphics[width=0.19\textwidth]{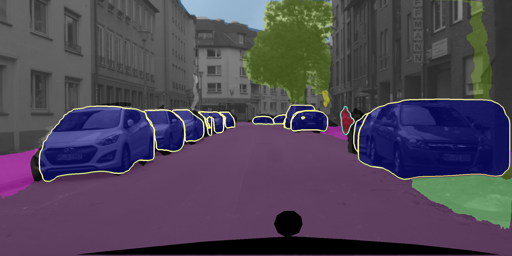}& \includegraphics[width=0.19\textwidth]{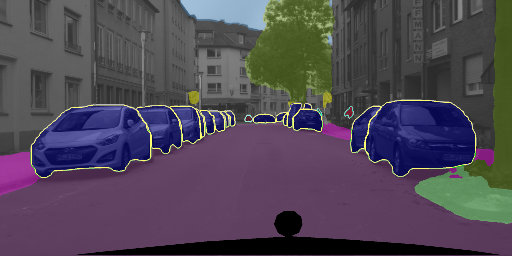}\\
         \includegraphics[width=0.19\textwidth]{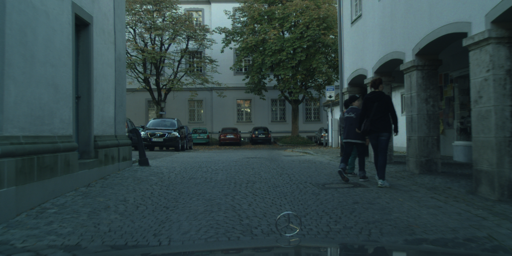}&
         \includegraphics[width=0.19\textwidth]{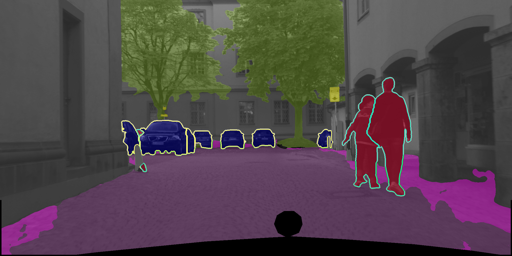}& \includegraphics[width=0.19\textwidth]{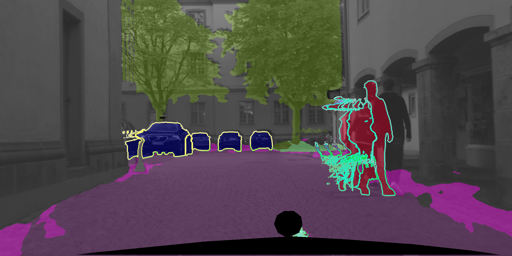}& \includegraphics[width=0.19\textwidth]{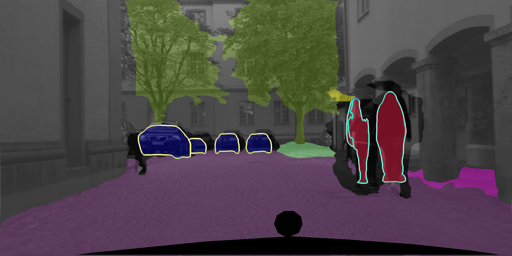}& \includegraphics[width=0.19\textwidth]{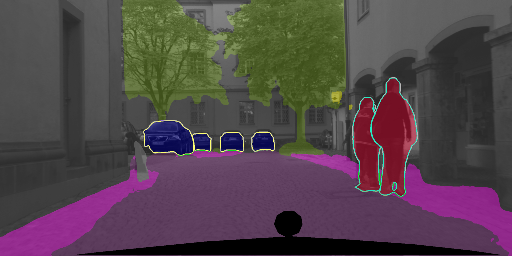}\\
    \end{tabular}
    \caption{\textbf{Additional mid-term panoptic segmentation forecasts on Cityscapes.}}
    \label{fig:panoptic_viz_mid_2}
    \vspace{-0.4cm}
\end{figure*}

\begin{figure*}[t]
    \centering
    \footnotesize
    \setlength\tabcolsep{0.5pt}
    \renewcommand{\arraystretch}{0.5}
    \begin{tabular}{ccccc}
        Last Seen Image &  Oracle & Flow & Hybrid & Ours\\
         \includegraphics[width=0.19\textwidth]{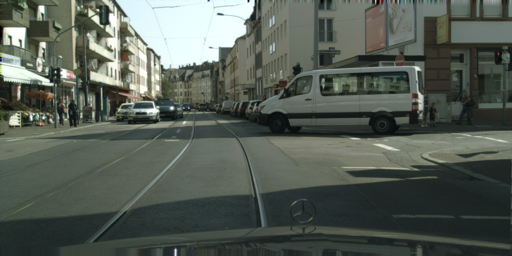}&
         \includegraphics[width=0.19\textwidth]{figs/viz/frankfurt_000000_002196_oracle.png}& \includegraphics[width=0.19\textwidth]{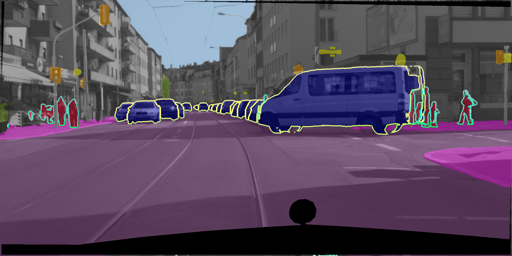}& \includegraphics[width=0.19\textwidth]{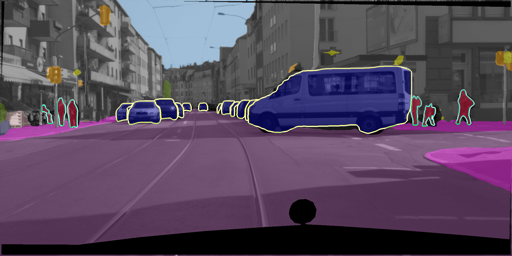}& \includegraphics[width=0.19\textwidth]{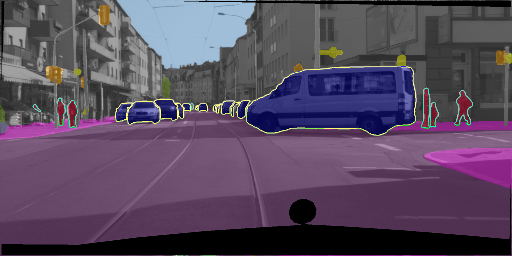}\\
         \includegraphics[width=0.19\textwidth]{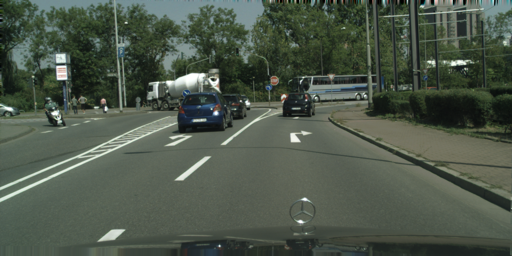}&
         \includegraphics[width=0.19\textwidth]{figs/viz/frankfurt_000000_005898_oracle.png}& \includegraphics[width=0.19\textwidth]{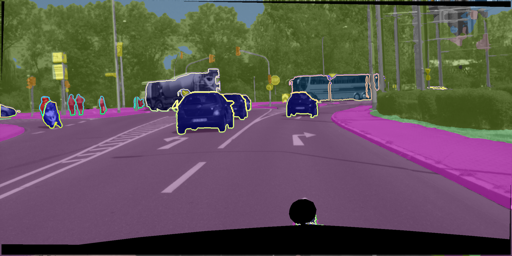}& \includegraphics[width=0.19\textwidth]{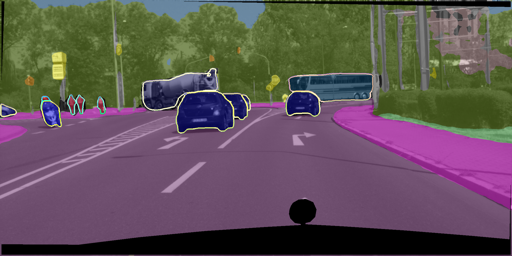}& \includegraphics[width=0.19\textwidth]{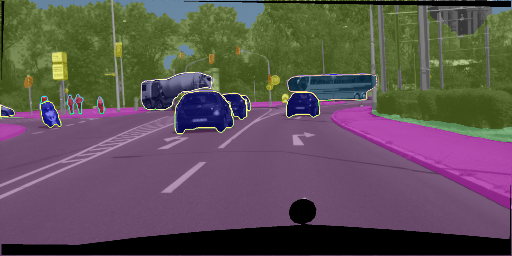}\\
         \includegraphics[width=0.19\textwidth]{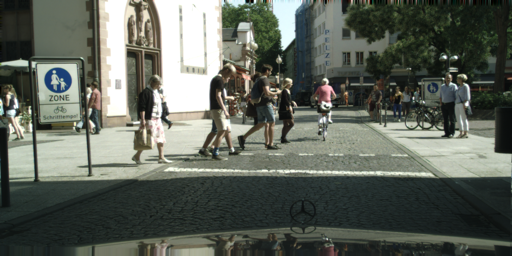}&
         \includegraphics[width=0.19\textwidth]{figs/viz/frankfurt_000001_011715_oracle.png}& \includegraphics[width=0.19\textwidth]{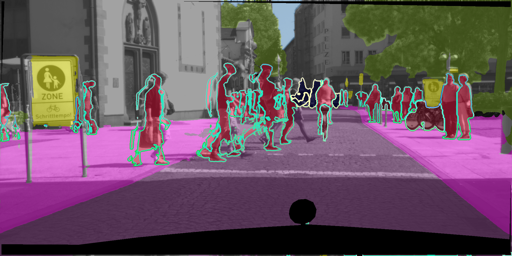}& \includegraphics[width=0.19\textwidth]{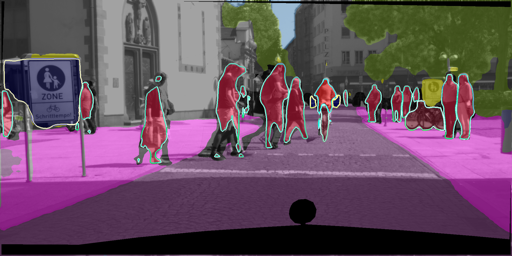}& \includegraphics[width=0.19\textwidth]{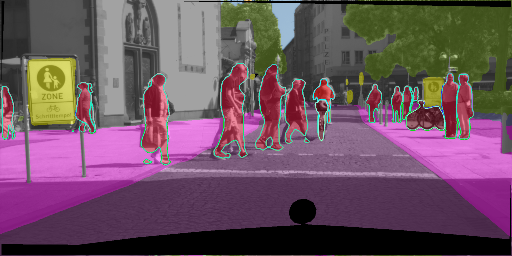}\\
         \includegraphics[width=0.19\textwidth]{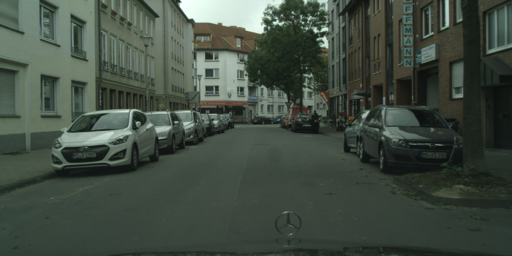}&
         \includegraphics[width=0.19\textwidth]{figs/viz/munster_000088_000019_oracle.png}& \includegraphics[width=0.19\textwidth]{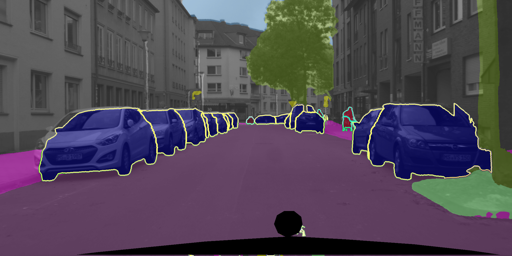}& \includegraphics[width=0.19\textwidth]{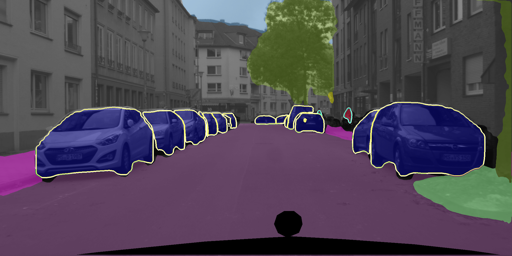}& \includegraphics[width=0.19\textwidth]{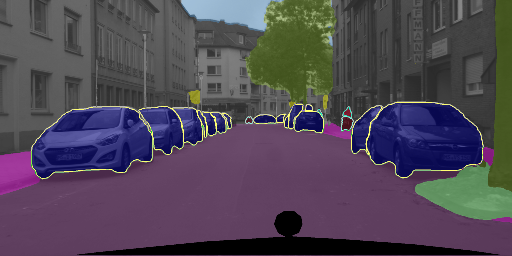}\\
         \includegraphics[width=0.19\textwidth]{figs/viz/lindau_000025_000016_leftImg8bit.png}&
         \includegraphics[width=0.19\textwidth]{figs/viz/lindau_000025_000019_oracle.png}& \includegraphics[width=0.19\textwidth]{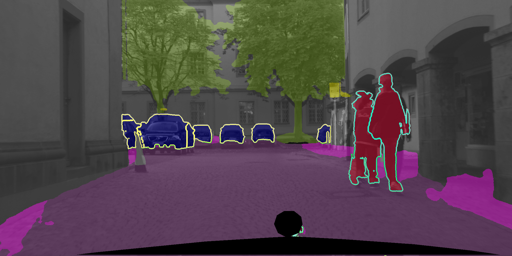}& \includegraphics[width=0.19\textwidth]{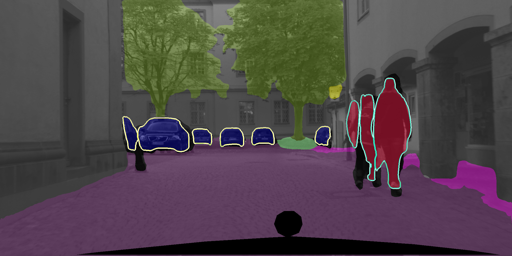}& \includegraphics[width=0.19\textwidth]{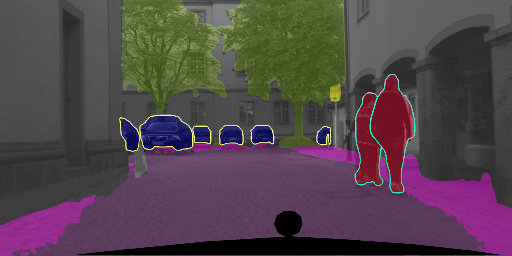}\\
    \end{tabular}
    \caption{\textbf{Additional short-term panoptic segmentation forecasts on Cityscapes.}}
    \label{fig:panoptic_viz_short_2}
    \vspace{-0.4cm}
\end{figure*}

\figref{fig:panoptic_viz_short} presents visualizations for the short term setting for the sequences present in \figref{fig:panoptic_viz}. Additional sequences are visualized for the mid-term setting in \figref{fig:panoptic_viz_mid_2}, and the corresponding short term sequences are presented in \figref{fig:panoptic_viz_short_2}.

Included with the supplementary material are some videos visualizing predictions. Each displays the most recently seen input frame in the top half and the prediction $9$ frames in the future from the last seen frame (\ie, for the mid-term setting). The folder `\texttt{short\_videos}' contains visualizations for some of the sequences present within the Cityscapes validation dataset. The folder `\texttt{long\_videos}' contains visualizations for a longer sequence taken from the unfiltered Frankfurt sequence provided with Cityscapes.

\end{document}